\documentclass{article}

\usepackage{arxiv}

\usepackage[utf8]{inputenc} 
\usepackage[T1]{fontenc}    
\usepackage{hyperref}       
\usepackage{url}            
\usepackage{booktabs}       
\usepackage{amsfonts}       
\usepackage{nicefrac}       
\usepackage{microtype}      
\usepackage{graphicx}
\usepackage{natbib}
\usepackage{doi}

\usepackage{algorithm}
\usepackage{algorithmic}
\usepackage{xcolor,colortbl}
\usepackage{adjustbox}
\usepackage{tabularx}
\usepackage{float}
\usepackage{multirow}
\usepackage{booktabs}
\usepackage{caption}
\usepackage{subcaption}
\usepackage{amsmath}
\usepackage{pdflscape}

\title{Using 3-Objective Evolutionary Algorithms for the Dynamic Chance Constrained Knapsack Problem}

\date{} 					

\author{ 
{Ishara Hewa Pathiranage}\\
  Optimisation and Logistics, \\School of Computer and Mathematical Sciences,\\
  The University of Adelaide,\\
  Adelaide, Australia \\
   \And
 {Frank Neumann}\\
  Optimisation and Logistics, \\School of Computer and Mathematical Sciences,\\
  The University of Adelaide,\\
  Adelaide, Australia \\
  \And
 {Denis Antipov}\\
  Optimisation and Logistics, \\School of Computer and Mathematical Sciences,\\
  The University of Adelaide,\\
  Adelaide, Australia \\
  \And
 {Aneta Neumann}\\
  Optimisation and Logistics, \\School of Computer and Mathematical Sciences,\\
  The University of Adelaide,\\
  Adelaide, Australia \\
  \And
}

\renewcommand{\shorttitle}{Using 3-Objective Evolutionary Algorithms for the DCCKP}

\hypersetup{
pdftitle={Using 3-Objective Evolutionary Algorithms for the Dynamic Chance Constrained Knapsack Problem},
pdfsubject={cs.NE},
pdfauthor={Ishara Hewa Pathiranage, Frank Neumann, Denis Antipov, Aneta Neumann},
pdfkeywords={Fitness evaluation, Combinatorial optimization, Multi-objective optimization, Constraint handling, Dynamic optimization, Chance constraints},
}

\begin{document}
\maketitle

\begin{abstract}
    Real-world optimization problems often involve stochastic and dynamic components. Evolutionary algorithms are particularly effective in these scenarios, as they can easily adapt to uncertain and changing environments but often uncertainty and dynamic changes are studied in isolation. In this paper, we explore the use of 3-objective evolutionary algorithms for the chance constrained knapsack problem with dynamic constraints. In our setting, the weights of the items are stochastic and the knapsack's capacity changes over time. We introduce a 3-objective formulation that is able to deal with the stochastic and dynamic components at the same time and is independent of the confidence level required for the constraint. This new approach is then compared to the 2-objective formulation which is limited to a single confidence level. We evaluate the approach using two different multi-objective evolutionary algorithms~(MOEAs), namely the global simple evolutionary multi-objective optimizer~(GSEMO) and the multi-objective evolutionary algorithm based on decomposition~(MOEA/D), across various benchmark scenarios. Our analysis highlights the advantages of the 3-objective formulation over the 2-objective formulation in addressing the dynamic chance constrained knapsack problem.
\end{abstract}


\section{Introduction}

Evolutionary algorithms~(EAs) have been widely applied to address a wide range of real-world combinatorial optimization problems, as these techniques proved themselves to be effective at providing robust solutions to complex problems with minimal design effort~\cite{DBLP:conf/evoW/2023cop}. In recent years, these techniques have been used to solve complex real-world problems across various domains, including mining, power systems, healthcare, and communication systems, to enhance productivity and reduce operational costs~\cite{article1, reid2021advanced, DECERLE2019712, GENG2019341, ABE20203304}.

These problems often involve stochastic as well as dynamic components in practice. Many real-world problems involve uncertainty with the nature of data, measurement errors, etc., and ignoring these stochastic components may lead to sub-optimal solutions or the computed optimal solutions might appear infeasible in practice. Furthermore, the objective function, constraints, or decision variables of optimization problems can be dynamic since these components may change over time. Therefore, when optimizing, the challenge is to track the movements of optima induced by dynamic changes~\cite{10.1145/3524495}. EAs have been successfully applied to solve many stochastic and dynamic problems due to their ability to easily adapt to changing environments~\cite{RAKSHIT201718, Neumann2020, Roostapour2020, DBLP:conf/ecai/AssimiHXN020,NGUYEN20121}. The development of methods for incorporating chance constraints into optimization problems allowed to use EAs for solving chance constrained problems with any specified probability denoted as $\alpha$~\cite{LI20151156} of satisfying the constraints.

In recent years, there has been a notable shift towards employing multi-objective formulations within EAs to address stochastic and dynamic problems. This approach, leveraging the ability to optimize multiple conflicting objectives simultaneously, offers a more holistic and realistic framework for solving real-world problems. This multi-objective optimization leads to obtaining a set of non-dominated solutions instead of a single solution.

However, there are only a few studies targeting both stochastic and dynamic components with multi-objective approaches. The dynamic chance constrained knapsack problem~(DCCKP) is investigated in ~\cite{DBLP:conf/ecai/AssimiHXN020} using evolutionary 2-objective optimization. But this approach is limited to single confidence level. It has been shown in ~\cite{DBLP:conf/gecco/0001W23} that 3-objective Pareto optimization is beneficial to tackle chance constrained problems for any confidence level $\alpha$ at once. In this work, we investigate the 3-objective EA approach to solve the DCCKP and its ability to cope with the stochastic and dynamic nature of this optimization problem together, independently of the confidence level imposed on the chance constraint.  

\subsection{Related Work}

EAs have been successfully applied to chance constrained optimization problems in the literature~\cite{POOJARI20081128, 10.1109/TEVC.2013.2244898, Xie2019,DBLP:conf/gecco/XieN020,DBLP:conf/gecco/XieN0S21,DBLP:conf/gecco/NeumannB021}. \citet{POOJARI20081128} proposed a framework combining genetic algorithm and Monte Carlo simulation to solve chance-constraint programming~(CCP) problems. The genetic algorithm is used to cope with the complicated landscape of CCP problems, while the simulation addresses the randomness in constraints. \citet{Xie2019} explored the integrating tail inequalities, such as Chebyshev's and Chernoff, to address the chance constrained knapsack problem, where the weight of each item is uniformly distributed. They evaluated performance of EAs on different knapsack instances, covering a wide range of stochastic settings. The authors of~\cite{DBLP:conf/ijcai/0001W22} conducted a runtime analysis of EAs on chance constrained optimization problems with independent and normally distributed stochastic components, using a 2-objective formulation. This research was further extended in~\cite{DBLP:conf/gecco/0001W23}. They theoretically investigated a 3-objective formulation, considering the deterministic constraint as the third objective with MOEAs for solving the chance constraint optimization problems. The results indicate that the 3-objective setting is particularly advantageous for graphs of moderate size in various stochastic settings.

EAs also performed well on deterministic dynamic problems. \citet{Roostapour2020} investigated both single and multi-objective baseline evolutionary algorithms in the context of the dynamic classical knapsack problem where the knapsack capacity varies over time. Their findings suggest that the multi-objective approach generally outperforms the single-objective algorithm, when the frequency of dynamic changes is not too high. Moreover, they stated that their simple multi-objective method yields better results than well-established evolutionary multi-objective algorithms like NSGA-II and SPEA2. \citet{Roostapour_Neumann_Neumann_Friedrich_2019} investigated the class of submodular functions with dynamically changing constraint bounds. It was shown that the simple multi-objective evolutionary approach, named POMC, effectively ensures the same worst-case approximation ratio in dynamic environments as that of classical greedy algorithms in static settings. \citet{10.1145/3321707.3321792} analyzed the performance of simple evolutionary algorithms theoretically for dynamic combinatorial optimization problems focusing the classical graph coloring problem. They introduced the dynamic setting by adding edges to the current graph. They showed that the re-optimization is much more efficient than optimizing from scratch for dynamic problems.

Even though there is growing interest in applying EAs to either dynamic or stochastic optimization problems, many real-world problems often involve stochastic elements~\cite{DBLP:conf/ppsn/NeumannXN22,E23OMOEA,TCHTTP,POCCMSP,ERDCCCMCP} or both stochastic and dynamic elements~\cite{DBLP:conf/ecai/AssimiHXN020,MOEAwSWS}. However, only few studies have focused on addressing both of them at the same time. In~\cite{DBLP:conf/ecai/AssimiHXN020}, a novel approach for solving the DCCKP was proposed. They introduced a 2-objective formulation to cater both uncertain and dynamic nature of the knapsack problem where the weight of each item is independent and uniformly distributed. The stochastic bound was used as the second objective. They estimated the probabilistic constraint using prominent tail inequalities Chebyshev's inequality and the Chernoff bound. The study evaluated the effectiveness of both single- and multi-objective EAs, finding that the 2-objective optimization is particularly effective in managing dynamic chance constrained problems. But their method is limited to one confidence level at once, which implies that it is required to find trade offs with respect to each confidence level separately.

\subsection{Our contribution}
In this paper, we consider the dynamic chance constrained knapsack problem, characterized by items with stochastic weights that are independent and normally distributed, while the profits remain deterministic. Such weights distribution allows the exact reformulation of the constraint, eliminating the need for tail bounds or sampling-based methods to assess solution feasibility. It's important to note, however, that this problem is still NP-hard and involves non-linear constraints. Additionally, in our dynamic setting, the knapsack capacity changes over time. We refer to this problem as the DCCKP in the rest of the paper. 

We introduce a 3-objective formulation to solve this DCCKP and evaluate its ability to cope with the stochastic and dynamic nature. This 3-objective formulation trade-offs the expected value and variance of the solution's weight, addressing different components of uncertainty along with the original objective function. Significantly, this reformulation provides optimal solutions for any linear combination of the expected value and standard deviation of stochastic items and addresses the original chance constrained problem for any confidence level $\alpha \geq 1/2$. This method eliminates the need to predefine a confidence level when applying multi-objective evolutionary algorithms and, by introducing an additional objective, offers a novel strategy to find high-quality solutions across a range of $\alpha$ values at once. Then we compare our approach with the 2-objective formulation motivated by the recent investigation in~\cite{DBLP:conf/ecai/AssimiHXN020} which focused on the original objective function, incorporating the reformulated exact constraints for a specific confidence level $\alpha$ on two multi-objective evolutionary algorithms, namely the GSEMO and the MOEA/D with different decomposition methods.



The rest of the paper is structured as follows. In Section~\ref{Sec: Problem_Algo}, we introduce the DCCKP and the baseline evolutionary algorithms that we used to evaluate the performance of our approach. Then in Section~\ref{Sec: 2-obj} and~\ref{Sec: 3-obj}, we present  2-objective and 3-objective formulations that we introduce for our investigation respectively. Then, we describe the experimental investigations in detail in Section~\ref{Sec: Investigations}. Finally, we conclude with some key remarks and insights in Section~\ref{Sec: Conclusion}.

\section{Preliminaries}
In this section, we define the DCCKP and we discuss the algorithmic setup. 
\label{Sec: Problem_Algo}
\subsection{Problem Definition}
In the classical knapsack problem, we have a set of $n$ items, each with its profit $p_i$ and weight $w_i$, and a knapsack with a capacity constraint $B$. A solution, denoted as $x$, is a bit string of length $n$, where each bit represents whether an item is included or not. The total weight of a solution is $w(x) = \sum_{i=1}^n w_i x_i$, and its profit is $p(x) = \sum_{i=1}^n p_i x_i$. The objective of this classical knapsack problem is to select a combination of items that maximizes the total profit while satisfying the weight bound constraint.

In this paper, we consider the optimization of a deterministic objective function $p(x)$ under a chance constraint. The knapsack capacity $B$ dynamically changes during the optimization. The weight of each item, $w_i$ follows the normal distribution $\mathcal{N}(\mu_i,\,\sigma_i^{2})$, $1\leq i\leq n$ with expected value $\mu_i$ and variance $\sigma_i^2$ and is independent of each other. For a given solution $x$, let $p(x)$ be the objective function value defined by the total profit, $w(x)$ be the weight of this solution (which is a random variable), $\mu(x) = \sum_{i=1}^n \mu_i x_i$ be the expected weight, and $v(x)= \sum_{i=1}^n \sigma_i^2 x_i$ be the variance of the weight.

The chance constrained knapsack problem is then formulated as
\begin{eqnarray}
\textbf{Maximize} & p(x)=\sum_{i=1}^n p_i x_i \label{eq: problem_s_1}\\
\textbf{Subject to} & \Pr(w(x) \leq B) \geq \alpha.\label{eq: problem_s_2}
\end{eqnarray}
The goal is to find a solution with the maximum profit such that the chance constraint, where the weight bound constraint should be satisfied with high probability $\alpha$.  We assume that $\alpha \in [\alpha_l, \alpha_h]$ where $\alpha_l > 1/2$ and $\alpha_h<1$, focusing on solutions with high confidence.
\begin{figure}
    \centering
    \begin{subfigure}[b]{0.45\textwidth}
        \centering
        \includegraphics[scale=0.45]{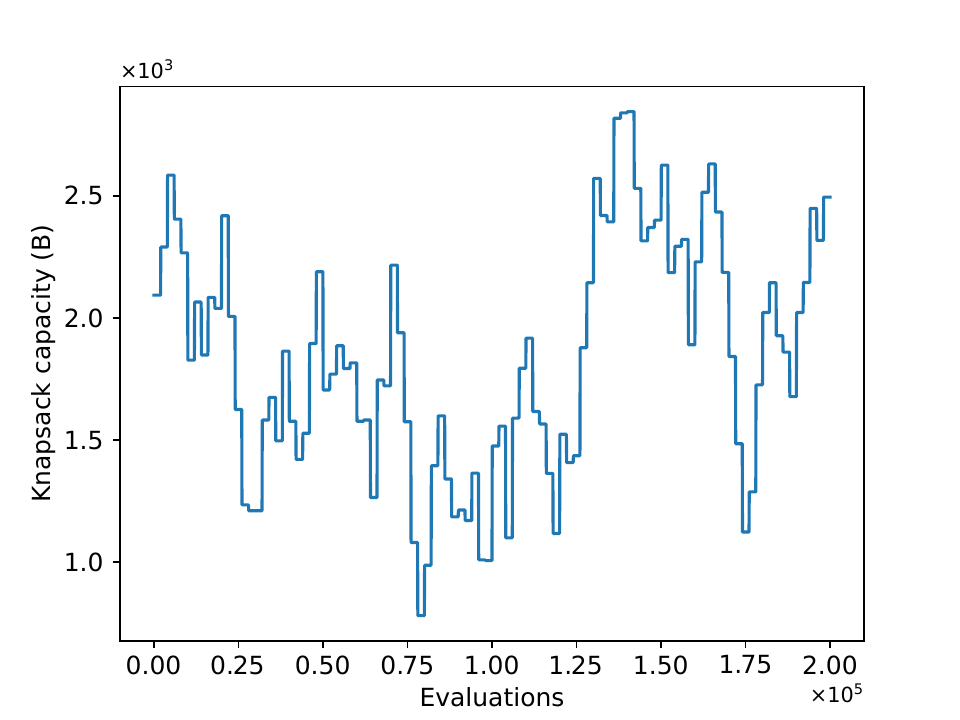}
        \caption{r=500}
        \label{fig:bound_r_500}
    \end{subfigure}
    \begin{subfigure}[b]{0.45\textwidth}
        \centering
        \includegraphics[scale=0.45]{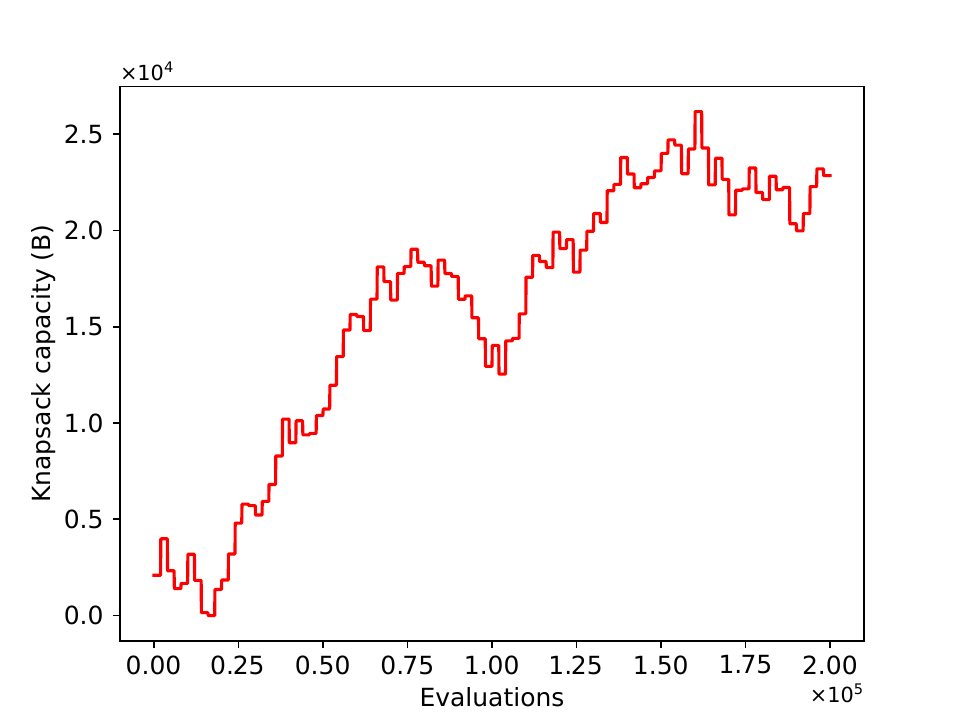}
        \caption{r=2000}
        \label{fig:bound_r_2000}
    \end{subfigure}
    
    \caption{Examples of the knapsack capacity changing over time for low ($r = 500$) and high ($r = 2000$) magnitude of changes.}
    \label{fig:bound}
\end{figure}
In the dynamic setting, the knapsack capacity $B$ changes over time every $t$ iterations. We assume that $t$ denotes the number of evaluations in between a knapsack capacity changes with magnitude $r$ according to the uniform distribution ~$\mathcal{U}(-r, r)$. The algorithm has $t$ generations to find the optimum of the current capacity and to prepare for the next change. Figure \ref{fig:bound} illustrates how dynamic changes affects to the Knapsack capacity. With the changes under $\mathcal{U}(-500, 500)$, the capacity reaches almost $3000$ within $100$ changes~(Figure~\ref{fig:bound_r_500}). While the maximum capacity increases up to $25000$ under $\mathcal{U}(-2000, 2000)$~(Figure~\ref{fig:bound_r_2000}). This implies that the algorithms have to consider various challenges that arise due to the dynamic nature of the problem.

\subsection{Baseline Evolutionary Algorithms}
A multi-objective optimization problem is defined as follows:
\begin{equation}\label{eq: mop}
    \textbf{Maximize}~F(x) = (f_1(x)\,\dots\,f_m(x))~\textbf{subject to}~x\in \Omega,
\end{equation} 
where $\Omega$ is the variable space, $F \colon \Omega \rightarrow R^{m}$ has $m$ real-valued objective functions and $R^m$ is the objective space. In most of the time, objectives in Equation~\eqref{eq: mop} contradicts each other, therefore we consider the Pareto optimality among each objective considering the best tradeoffs among all the objectives. Let $x,y \in R^m$ and assume that we maximize all the objectives $f_i$, $i \in \{1\,\dots\,m\}$, then a solution ${x}$ dominates a solution ${y}$ (denoted by ${x} \succeq {y}$) iff $x_i\geq y_i$ for every $i$. Furthermore, a solution ${x}$ strongly dominates ${y}$ (denoted by ${x} \succ {y}$) iff ${x} \succeq {y}$ and $x_i > y_i$ for at least one index $i$.  

We employ multi-objective EAs to solve the DCCKP and this paper focuses on two specific algorithms, namely the GSEMO and the MOEA/D.
        
The GSEMO is the most basic evolutionary algorithm that addresses multi-objective optimization problems~\cite{gsemo}. Previously, it was proven to be effective in solving chance constrained multi-objective optimization problems in many studies~\cite{DBLP:conf/gecco/0001W23}. Its population $S$ initially contains a solution that is generated randomly. In each iteration, a parent solution $x$ is chosen uniformly at random from $S$. An offspring solution $y$ is then created by flipping each bit of $x$ with a probability of $1/n$. The offspring $y$ is added to the population $S$ if it is not strongly dominated by any existing solutions in $S$. Additionally, if $y$ is added, all solutions in $S$ that are dominated by $y$ are removed. This approach ensures that at the end of any iteration population $S$ will comprise a set of non-dominated solutions, each of which is equally efficient in terms of the given objective functions. For more details on the GSEMO please refer to~\cite{gsemo}.

Another algorithm we consider here is the MOEA/D~\cite{li2008multiobjective, 4358754} which decomposes the multi-objective optimization problem into several single-objective sub-problems, which are then optimized simultaneously. The objective of each sub-problem is an aggregate of all objectives, and neighborhood relations among these sub-problems are defined based on the distances between their aggregation coefficient vectors. This algorithm utilizes various decomposition approaches, such as Weighted Sum (WS), Tchebycheff, and Penalty Boundary Intersection (PBI)~\cite{li2008multiobjective, 4358754}, to transform a multi-objective problem into scalar sub-problems. For more details on the MOEA/D please refer to~\cite{li2008multiobjective, 4358754}. 


The WS approach involves maximizing a convex combination of the objectives and the optimal solution to the scalar optimization problem is given by:
$$ 
\textbf{Maximize} \ g^{ws}(x\mid\mathbf{\lambda}) = \sum_{i=1}^m \lambda_i f_i(x) ~
\textbf{subject to} \ x \in \Omega,
$$
where $\mathbf{\lambda} = (\lambda_1,\ldots,\lambda_m)$ is a weight vector which satisfies the conditions $\lambda_i \geq 0$ for all $i = 1,\ldots,m$ and $\sum_{i=1}^m \lambda_i = 1$ where $m$ is the number of objective functions. We generate such weight vectors using the Dirichlet distribution.  
In the Tchebycheff decomposition the scalar optimization problem is defined as:
$$
\textbf{Minimize} \ g^{te}(x | \lambda,z^{*}) = \max_{1 \leq i \leq m} \{\lambda_i |f_i(x) - z_i^{*}|\} 
\textbf{ subject to} \ x \in \Omega,
$$
where $z^{*} = (z_1^{*},\ldots,z_m^{*})$ is the reference point such that $z_i^{*}$ is the best value of $f_i(x)$ encountered so far by the algorithm. 
The PBI method minimizes the following scalar optimization sub-problem combining the distance to the reference point and a penalty term as mentioned below.
$$
\textbf{Minimize} \ g^{pbi}(x | \lambda,z^{*}) = d_1 + \theta d_2 ~
\textbf{subject to} \ x \in \Omega,
$$
where $d_1 = \frac{|(z^{*}-F(x)) \cdot \lambda|}{\mid\mid \lambda \mid\mid}$ and $d_2 = | F(x) - (z^* - d_1\lambda)|$. Parameter $\theta > 0$ is a pre-set penalty parameter.

\section{Dynamic 2-objective Formulation}
\label{Sec: 2-obj}

We motivate our 2-objective formulation by the recent investigation in~\cite{DBLP:conf/ecai/AssimiHXN020}. In that work, the authors considered the DCKP where the weights of each item are random and follow the uniform distribution and the capacity bound $B$ (denoted in that paper by $C$) changes over time. Their goal was to maximize the profit $p(x)$ such that the probability that the weight of the solution $w(x)$ is at least as high as the capacity $B$ is at most $\alpha$~(that is, $\Pr[w(x)\geq B] \leq \alpha$ for some small $\alpha$, which is different from our formulation, where the constraint is $\Pr[w(x)\leq B] \geq \alpha$ for some $\alpha$ close to one). To solve this problem, they introduced a 2-objective formulation where the stochastic bound $B^*(x)$ is the  second objective and defined $B^*(x)$ as the smallest weight bound $B'$ such that the probability that the weight $w(x)$ is at least $B'$ is at most $\alpha$. They used Chebyshev's inequality and the Chernoff bounds to estimate an upper bound on the chance constraint $B^*(x)$. 

We also adopt the standard Pareto optimization approach where weight and profit are the two objectives. Our goal is to select a subset of items that maximizes profit while satisfying the given chance constraint such that a solution satisfies the knapsack bound $B$, with a probability at least some $\alpha$ which is close to one. 

We formulate the $2$-objective approach to our problem as follows:
$$
g_{2D}(x)=(f(x), \tilde w(x)),
$$
where our aim is to maximize $f(x)$~(profit component) and minimize $\tilde w(x)$~(weight component).

We assume that the weight of each item is independent and follow a normal distribution, therefore, the total weight also follows a normal distribution. Hence, we do not need to estimate the lower bound on $w_\alpha(x)$ using tail inequalities, but we can estimate it precisely via the $\alpha$-fractional point of the standard normal distribution, denoted by $K_\alpha$.
According to~\cite{Ishii1981StochasticST}, we can compute the chance constraint weight to be equivalent to
\begin{equation}
w_{\alpha}(x)= \mu(x) + K_{\alpha} \sqrt{v(x)}.
\label{eq: chance_c_weight}
\end{equation}

In the $2$-objective approach, we consider one fixed $\alpha$ time, as $w_{\alpha}(x)$ depends on $\alpha$. It is important to note that $K_{\alpha}$ does not have a finite value when $\alpha=1$, due to the infinite tail of the normal distribution. Our analysis considers the range of $\alpha$ values where $K_{\alpha}\geq0$.

In the dynamic setting, when varying the bound $B$, we cater for changes in the constraint bound by computing solutions that are within an $\eta$ range of the bound $B$. 
We keep only solutions $x$ that satisfy the condition 
$B -\eta \leq w_{\alpha}(x) \leq B+ \eta$, where $\eta$ determines how far away from bound $B$ can individuals which are stored in the population be. 
The algorithm prepares for the next dynamic changes by storing nearby solutions, even if they are infeasible~(that is, with $w_\alpha(x) > B$) as they may become feasible after the next change. 

For the 2-objective formulation we use the following fitness functions.
\begin{eqnarray}
    f(x) &=& \begin{cases}
    p(x) & \text{if } w_{\alpha}(x) \in [B - \eta, B + \eta], \\
    -e(x) & \text{otherwise},
\end{cases} 
\label{eq: objective_1_2d_d}\\
\tilde w(x) &=& \begin{cases}
   w_\alpha(x) & \text{if } w_{\alpha}(x) \in [B - \eta, B + \eta], \\
    w_{\alpha_{max}}+1 + e(x) & \text{otherwise}.
\end{cases}
\label{eq: objective_2_2d_d}
\end{eqnarray}
where $p(x) = \sum_{i=1}^n p_i x_i$ is the total profit of selected items, $n$ is the number of items, $w_{\alpha_{max}} = \mu_{max}\cdot n + K_{\alpha}\cdot\sqrt{v_{max}\cdot n}$ is the maximum possible value of $w_\alpha$ calculated based on the maximum expected weight, $\mu_{max} = \max_{1 \leq i \leq n}\mu_i$ and maximum variance, $v_{max} = \max_{1 \leq i \leq n}v_i$ among all the items. The penalty term $e(x)$ is calculated as $|w_{\alpha}(x)-B| - \eta$ indicating how far a solution $x$ is from the considered feasible range. 

All solutions are re-evaluated corresponding to the new bound after each dynamic magnitude change. If all current stored solutions fall outside  of the feasible range, namely $[B-\eta, B+\eta]$, then the algorithm considers the previous best solution~(the one with the maximum profit) as the initial solution and employs the repair mechanism (see Algorithm \ref{algo: three}), which behaves similar to the $(1+1)$ EA, until a feasible solution is found or termination criteria is met. In the GSEMO, after finding a feasible solution we start the optimization process using this solution and in the MOEA/D algorithm, we update the previously selected solution from the population with this feasible solution.

\begin{algorithm}[t!]
  \caption{Repair Mechanism}
  \label{algo: three}
  \begin{algorithmic}[1]
    \STATE $x \gets$ best previous solution;
    \WHILE{$x$ is not feasible}
      \STATE $y \gets$ flip each bit of $x$ independently with probability of $\frac{1}{n}$;
      \IF{$f_r(y) \geq f_r(x)$}
        \STATE $x \gets y$;
      \ENDIF
    \ENDWHILE
  \end{algorithmic}
\end{algorithm}

The fitness function that we use in the repair mechanism is:
\begin{equation}
\label{eq:repair}
f_r(x) = p(x) - {(n \cdot p_{max} + 1) \cdot h(x),}
\end{equation}

where $p_{max} = \max_{1 \leq i \leq n}p_i$ is the maximum profit, $h(x) = \max\{| w_{\alpha}(x)-B | - \eta,0\}$ is the constraint violation. The algorithm aims to maximize $f_r(x)$ which has two terms, total profit $p(x)$ and the penalty term $h(x)$ which guarantees that a feasible solution always has a better $f_r$ value than an infeasible solution. If solution is feasible, then $h(x)=0$, otherwise we assign large penalty to the objective of the solution. For 2-objective formulation, the constraint violation, $h(x)$ represents the deviation of the chance constraint weight from the knapsack bound.


\section{Dynamic 3-objective Formulation}
\label{Sec: 3-obj}
We introduce the 3-objective formulation of the DCCKP with normally distributed random variables. In the 2-objective formulation, we consider the Pareto optimization approach that compute the trade-offs with respect to the original objective functions and given chance constraint. In contrast to that, we consider different components determining the uncertainty of solutions, namely the expected value and variance of a solution. Therefore, this formulation employs a Pareto optimization approach, computing trade-offs with respect to the profit $p(x)$, expected weight $\mu(x)$, and variance of weight $v(x)$. 

In the 3-objective approach we use the objective function given as $g_{3D}(x)=(f_1(x), f_2(x), f_3(x))$ where 
\begin{align}
f_1(x) &= \begin{cases}
    p(x) &   \text{if $x$ feasible },  \\
    -e'(x) &   \text{otherwise},
\end{cases} 
\label{eq: objective_1_3d_d}\\
f_2(x) &= \begin{cases}
    \mu(x) & \text{if $x$ feasible},  \\
    n\cdot\mu_{max}+1 + e'(x) & \text{otherwise},
\end{cases}
\label{eq: objective_2_3d_d}\\
f_3(x) &= \begin{cases}
    v(x)   & \text{if $x$ feasible}, \\
    n\cdot v_{max}+1+e'(x) & \text{otherwise}.
\end{cases} 
\label{eq: objective_3_3d_d}
\end{align}
In this formulation, we maximize the profit $f(x)$ and minimize uncertainty components, $\mu(x)$ and $v(x)$ as two separate objectives in $f_2(x)$ and $f_3(x)$ instead of minimizing the chance constraint weight $w_{\alpha}(x)$ in the 2-objective formulation. This allows us to compute trade-offs that satisfy all confidence levels at once since $f_2(x)$ and $f_3(x)$  are independent of confidence level $\alpha$.

Solution $x$ is considered feasible if $x$ satisfies $(B-\eta \leq w_{\alpha_l}(x) \leq B+\eta) \vee 
(B-\eta \leq w_{\alpha_h}(x) \leq B+\eta) \vee
(w_{\alpha_l}(x) \leq B-\eta \wedge B+\eta \leq w_{\alpha_h}(x))$. Since $w_\alpha(x)$ is increasing in $\alpha$ for all $x$, this condition is equivalent to $[w_{\alpha_l}, w_{\alpha_h}] \cap [B-\eta, B+\eta] \ne \emptyset$. This indicates that the solution is feasible for at least a subset of confidence levels. Otherwise, the solution is considered as infeasible, and we assign a large penalty, $e'(x) = \max\{(B-\eta) - w_{\alpha_h}(x),w_{\alpha_l}(x) - (B+\eta)\}$. This implies that each objective value of an infeasible solution is always worse than the corresponding value of a feasible solution. 

After each magnitude change of knapsack bound, if there is no any feasible solution with respect to the new bound, then repair mechanism~(Algorithm~\ref{algo: three}) is performed and penalty term $h(x)$ in Equation~\eqref{eq:repair} is obtained as $
\max\{0, \max\{(B-\eta) - w_{\alpha_h}(x),0\} + \max\{w_{\alpha_l}(x) - (B+\eta),0\}\}$ which is the maximum weight bound violation with respect to considered $\alpha$ range. This approach effectively eliminates the selection of infeasible solutions.

\section{Experimental Investigations}
\label{Sec: Investigations}

In this section, we carry out the experimental investigations to see the effectiveness of the 3-objective formulation on the DCCKP on different benchmark settings and compare the performance with the 2-objective approach.

\subsection{Experimental Setting}
\label{Sec: Experimental_setting}

We use two distinct classes of knapsack weights and profits: uncorrelated~(uncorr) and bounded strongly correlated~(bsc) from the eli101 benchmarks for this study~\cite{benchmark}. For the uncorr class, weights and profits are randomly generated, while for the bsc class, weights are chosen uniformly at random, and the profits are set as the weight plus a fixed number. We test each type of problem for $300, 500$ and $1000$ number of items.

In our chance-constraint setting, the weights of the knapsack items $w_i$ are assumed to have a normal distribution $\mathcal{N}(\mu_i,\sigma_{i}^{2})$ with expected value $\mu_i$ and variance $v_i = \sigma_i^2$, $1 \leq i \leq n$. We choose small and large variance $v_i$ as an integer chosen independently and uniformly at random for each item $i$ either from $\mathcal{U}(1, \mu_i)$, or from $\mathcal{U}(\mu_i^2, 2\mu_i^2)$ and denote these distributions by $V_1$ and $V_2$ respectively. For experiments, we consider $\alpha \in \{1-10^{-2},1-10^{-4},1-10^{-6},1-10^{-8},1-10^{-10}\}$ hence $\alpha_l=1-10^{-2}$ and $\alpha_h=1-10^{-10}$. These $\alpha$ values represent the reliability of the final solution in ascending order. 

We compare the GSEMO, and the MOEA/D with three different decomposition approaches, namely weighted sum~(MOEA/D\textsubscript{ws}), Tchebycheff ~(MOEA/D\textsubscript{te}), and PBI ~(MOEA/D\textsubscript{pbi}) across various stochastic settings. To generate new offspring, standard bit flip mutation with mutation rate $1/n$ is used in both algorithms. Additionally, in the MOEA/D, we apply the uniform crossover with probability $0.8$ in the reproduction phase. In MOEA/D, the population size is set to $n$ for the 2-objective formulation and $2n$ for the 3-objective formulation, as the number of trade-offs increases with the number of objectives. Note that the algorithm considered in~\cite{DBLP:conf/ecai/AssimiHXN020} is similar to the algorithm that we used as 2-objective GSEMO in our comparison.

For each experimental run, we allocate a fitness evaluation budget of one million ($1$M). In the 3-objective formulation, this budget allows us to obtain results for each  setting for all $\alpha$ values in a single run. However, in the 2-objective formulation, we divide the $1$M budget across five runs for different $\alpha$ values.

In dynamic settings, we change the knapsack bound every $t$ generations, with $t$ set to $2000$, $4000$, and $20000$ for the 2-objective formulation, and $10000$, $20000$, and $100000$ for the 3-objective formulation, representing large, medium, and small frequencies of change, respectively. In both formulations, the number of dynamic changes $\nu$ remains the same at $100$, $50$, and $10$ correspondingly. The magnitude of each dynamic change is randomly generated following $\mathcal{U}({-r}, r)$, with $r$ set to $500$ and $2000$ to demonstrate small and large magnitude changes to capacity. To effectively address dynamic changes, we provide an initial value of $\eta$ which specifies how far from $B$ that individuals can be. When dynamic changes arise from $\mathcal{U}({-r}, r)$, $\eta$ is set to $r$, corresponding to the maximum capacity change.

To evaluate the performance of our algorithms in dynamic environment, we measure the partial offline error~\cite{three}, which indicates the difference between the profit of best solution obtained by the algorithm and the profit of the optimal solution for the deterministic knapsack problem $P(x^*)$ where the weight of each item is equal to the mean weight of our chance constrained problem. We employ dynamic programming to compute the optimal solution for the deterministic knapsack problem with the maximum profit $P(x^*)$, and we recompute $P(x^*)$ for every change in the capacity of the knapsack. This method, while exact, can be computationally intensive for large instances but serves as a valuable benchmark for heuristic algorithms. We obtain $x_i$, which is the best solution obtained by the algorithm in the last generation before the $i$-th change. If $x_i$ is feasible then the offline error $e_i$ is calculated as $P({x}^*)- p(x_i)$. Otherwise, $e_i = P({x}^*) + \min(w_{\alpha}(x_i)) - B$, where we add the minimum constraint violation to the maximum profit $P({x_i}^*)$. Finally, we calculate the total partial offline error $E$ as $ \frac{\sum_{j=1}^{\nu} e_j}{\nu}$. 

Statistical comparisons are carried out by using the Kruskal-Wallis test with 95\% confidence interval integrated with the Bonferroni post-hoc test to compare multiple solutions~\cite{Corder_Foreman_2014}. The tables representing the data obtained in our experiments together with the results of the statistical tests can be found in the Appendix. The stat shows the rank of each algorithm in the instances; If two algorithms can be compared with each other significantly, X \textsuperscript{(+)} denotes that the algorithm X is outperforming the current algorithm. Likewise, X\textsuperscript{(-)} implies the algorithm X is worse than the current algorithm significantly. X\textsuperscript{(*)} is denoted that there is no statistical significant between two algorithms.

\subsection{Experimental Results}
\label{Sec: Experimental_results}
\begin{figure*}
    \begin{subfigure}[b]{0.48\textwidth}
        \centering
        \includegraphics[width=\textwidth]{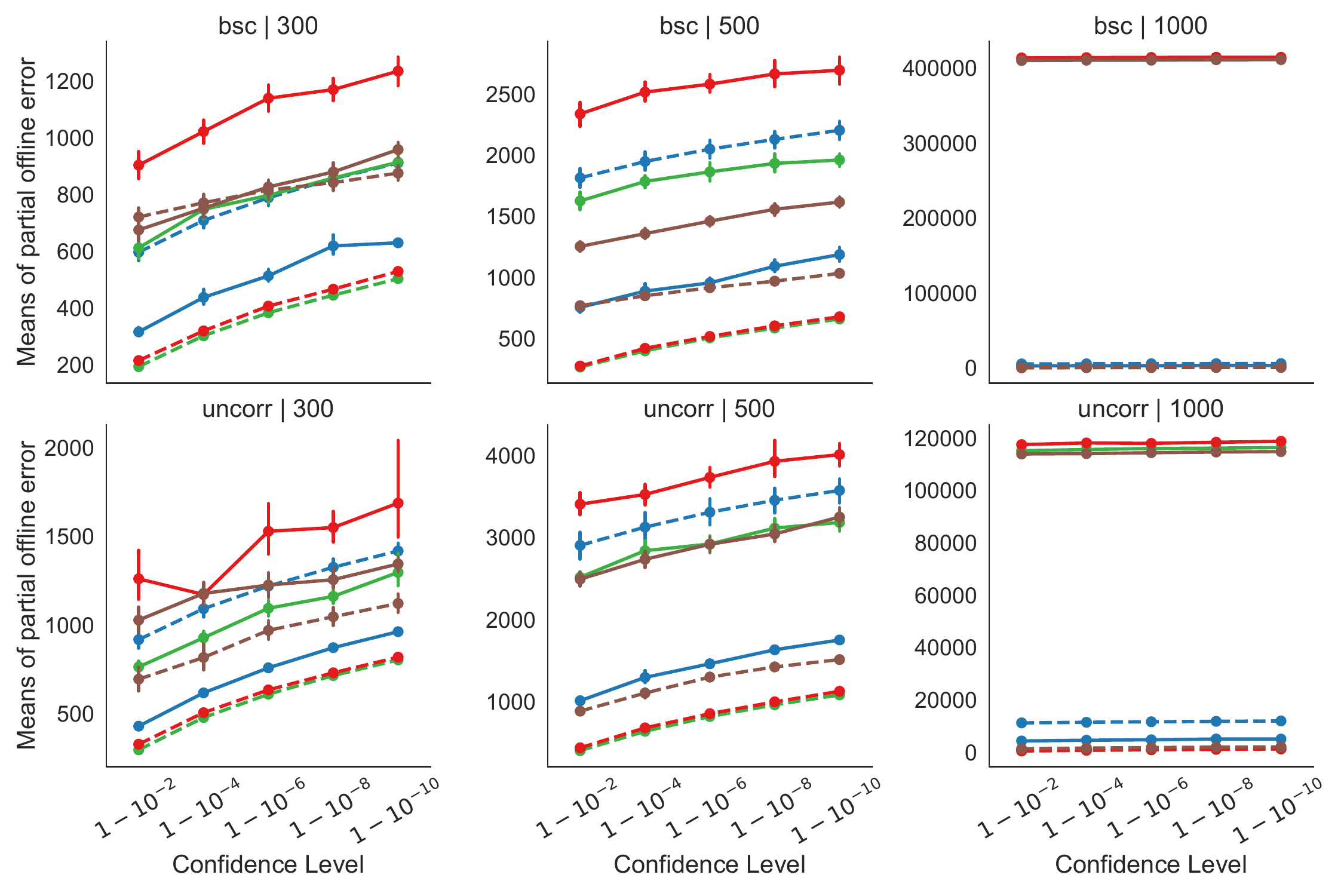}
        \caption{$\nu=10$ and the variance of each item is chosen according to $V_1$.}
        \label{fig: r_500_t_10_var1}
    \end{subfigure}
    \begin{subfigure}[b]{0.48\textwidth}
        \centering
    \includegraphics[width=\textwidth]{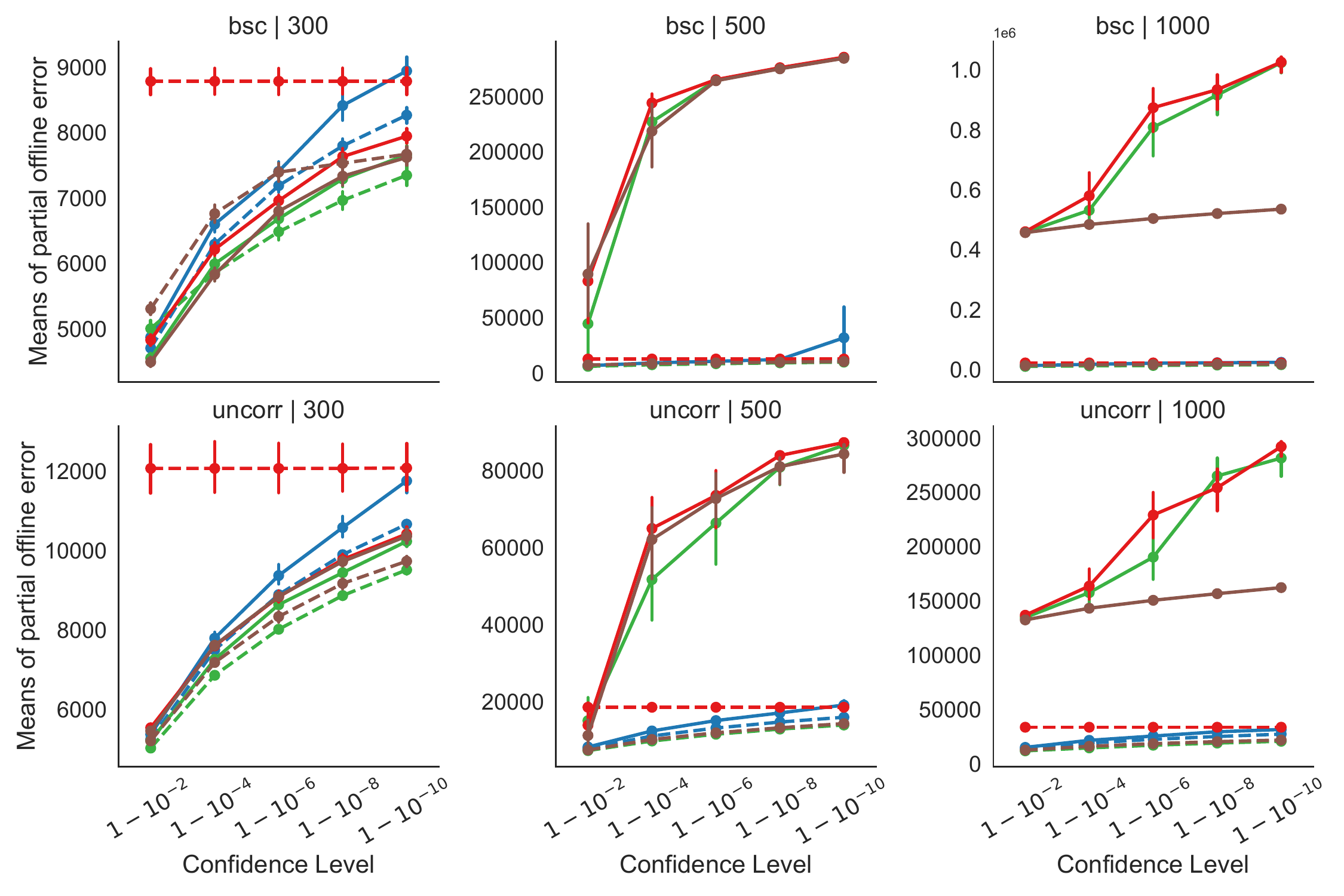}
        \caption{$\nu=10$ and the variance of each item is chosen according to $V_2$.}
        \label{fig: r_500_t_10_var2}
    \end{subfigure}
    
    \begin{subfigure}[b]{0.48\textwidth}
        \centering
        \includegraphics[width=\textwidth]{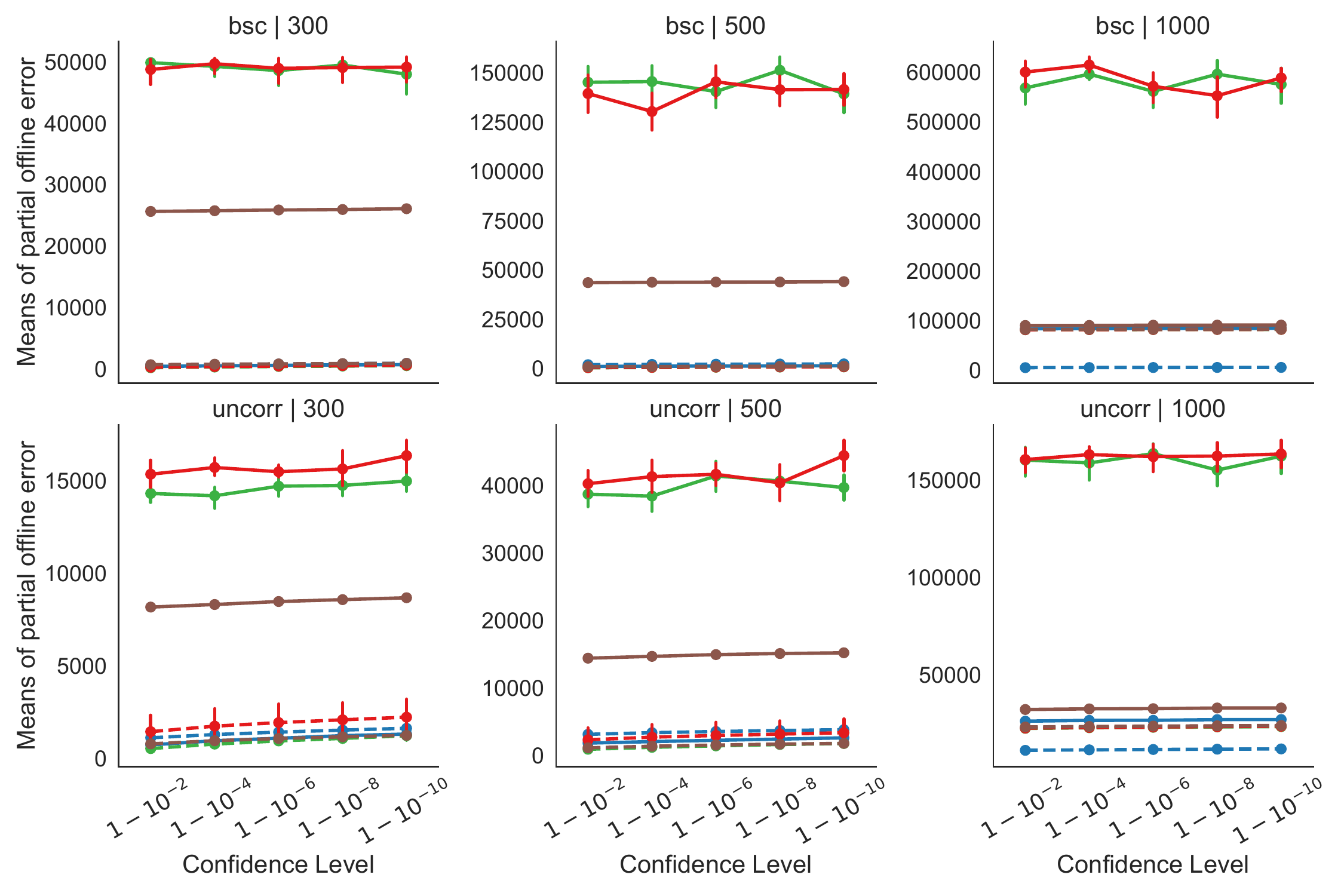}
        \caption{$\nu=50$ and the variance of each item is chosen according to $V_1$.}
        \label{fig: r_500_t_50_var1}
    \end{subfigure}
    \begin{subfigure}[b]{0.48\textwidth}
        \centering
        \includegraphics[width=\textwidth]{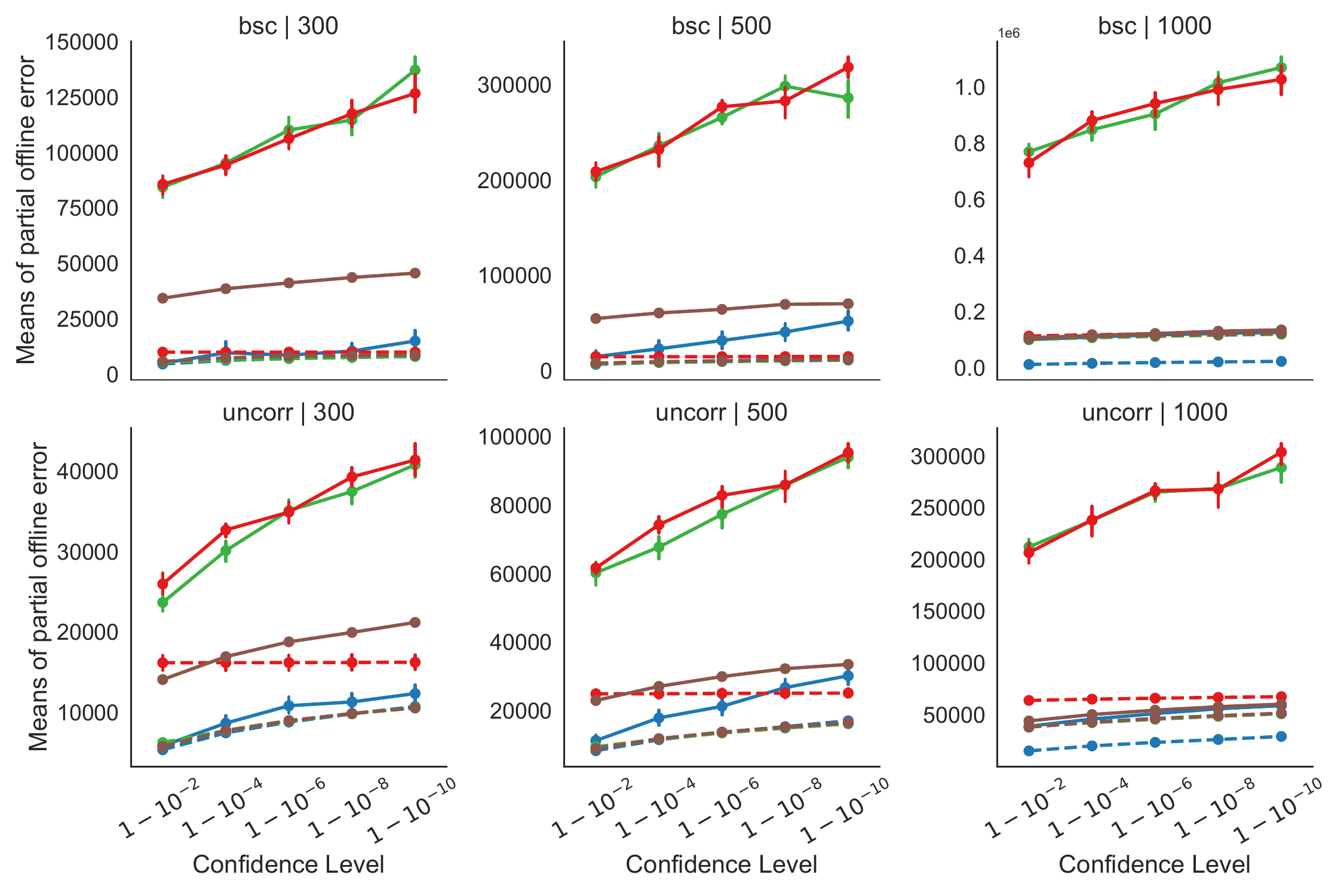}
        \caption{$\nu=50$ and the variance of each item is chosen according to $V_2$.}
        \label{fig: r_500_t_50_var2}
    \end{subfigure}

    \begin{subfigure}[b]{0.48\textwidth}
        \centering
        \includegraphics[width=\textwidth]{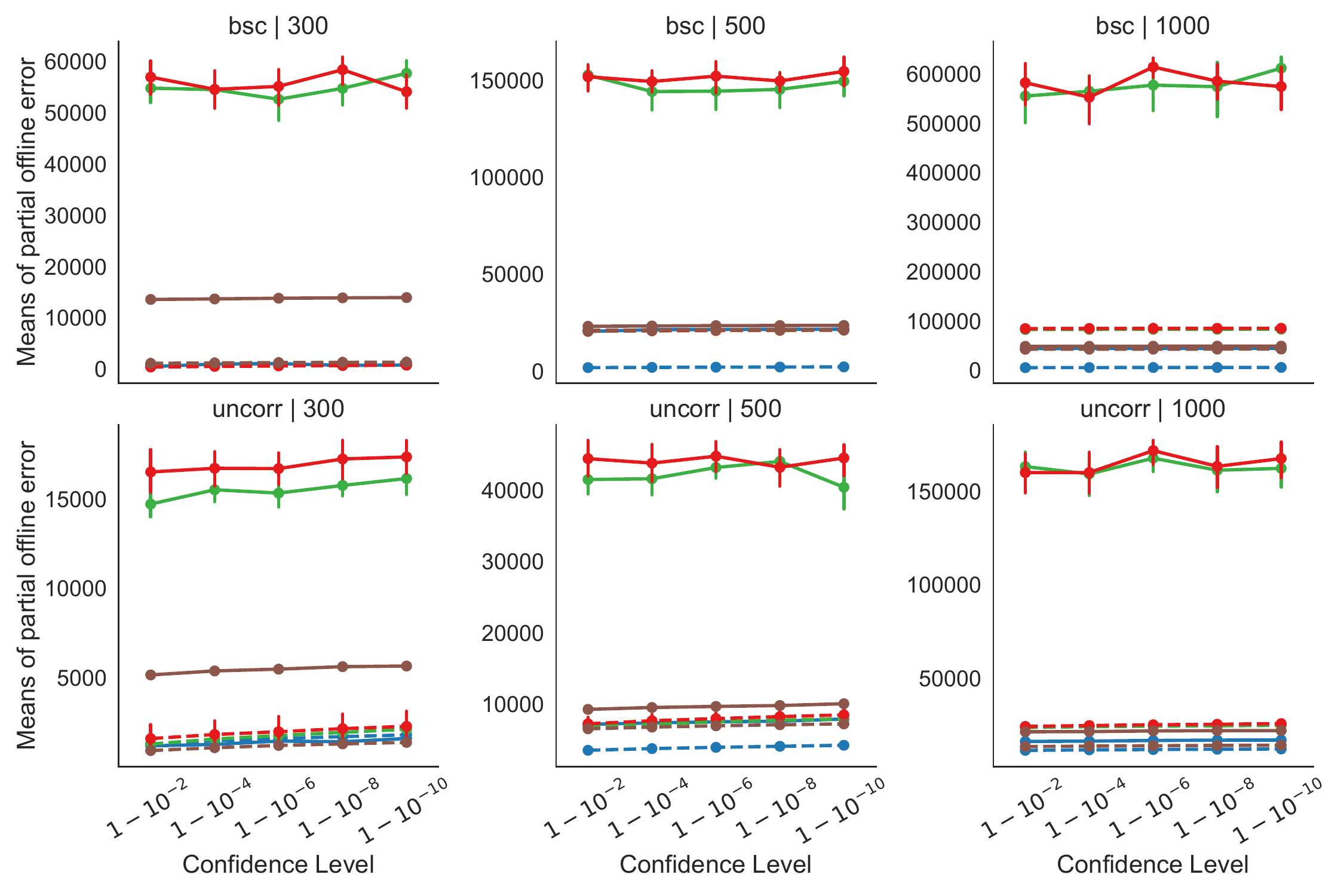}
        \caption{$\nu=100$ and the variance of each item is chosen according to $V_1$.}
        \label{fig: r_500_t_100_var1}
    \end{subfigure}
    \begin{subfigure}[b]{0.48\textwidth}
        \centering
        \includegraphics[width=\textwidth]{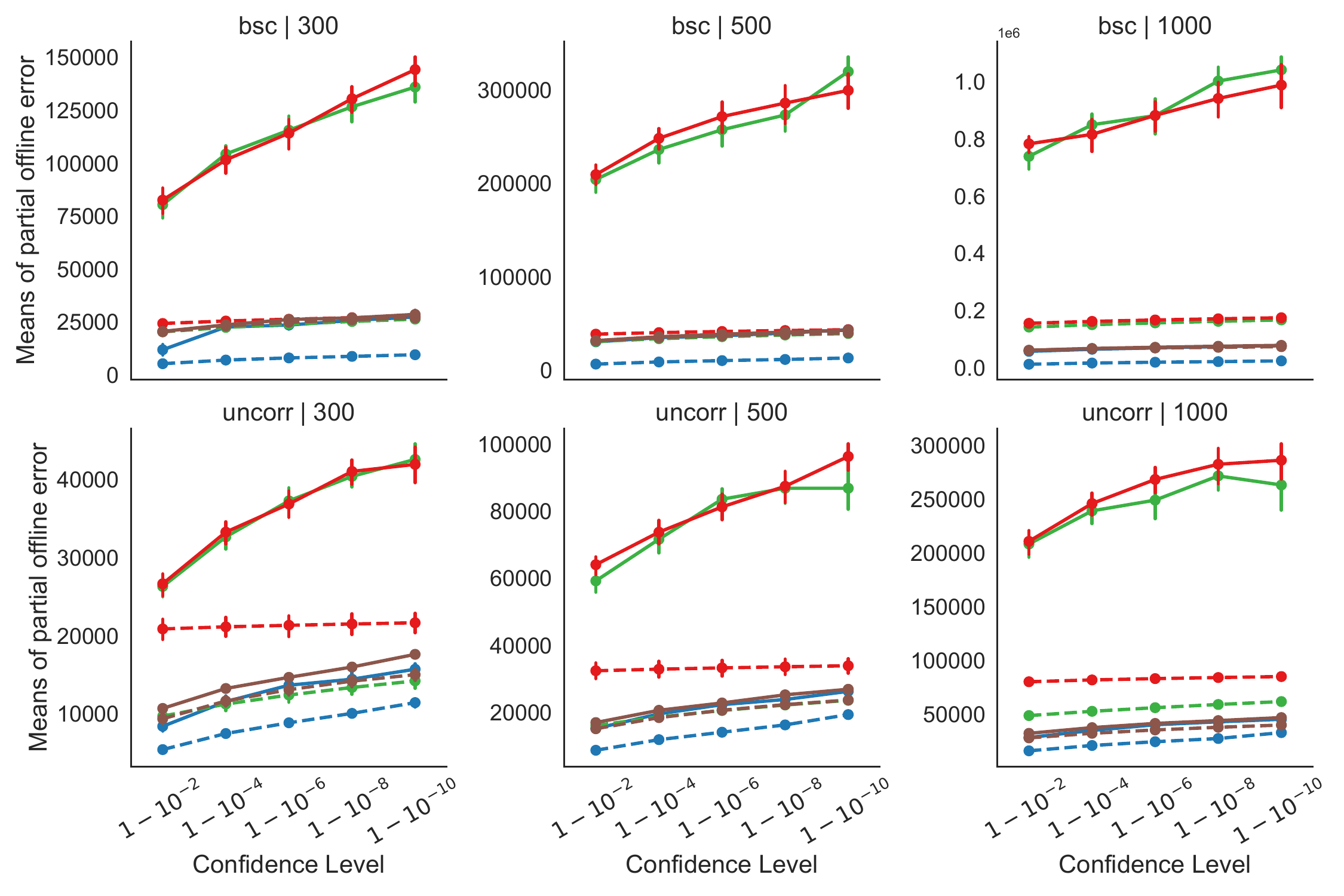}
        \caption{$\nu=100$ and the variance of each item is chosen according to $V_2$.}
        \label{fig: r_500_t_100_var2}
    \end{subfigure}
    \begin{subfigure}[b]{\textwidth}
        \centering
        \includegraphics[scale=0.4]{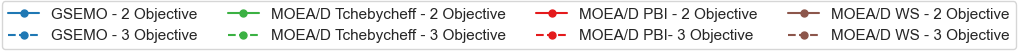}
    \end{subfigure}
    \caption{The mean value and the standard deviation of the total offline error for different algorithms when the magnitude of dynamic knapsack bound change follows $\mathcal{U}({-500}, 500)$ 
    }
    \label{fig: r_500}
\end{figure*}

\begin{figure*}
    \begin{subfigure}[b]{0.48\textwidth}
        \centering
        \includegraphics[width=\textwidth]{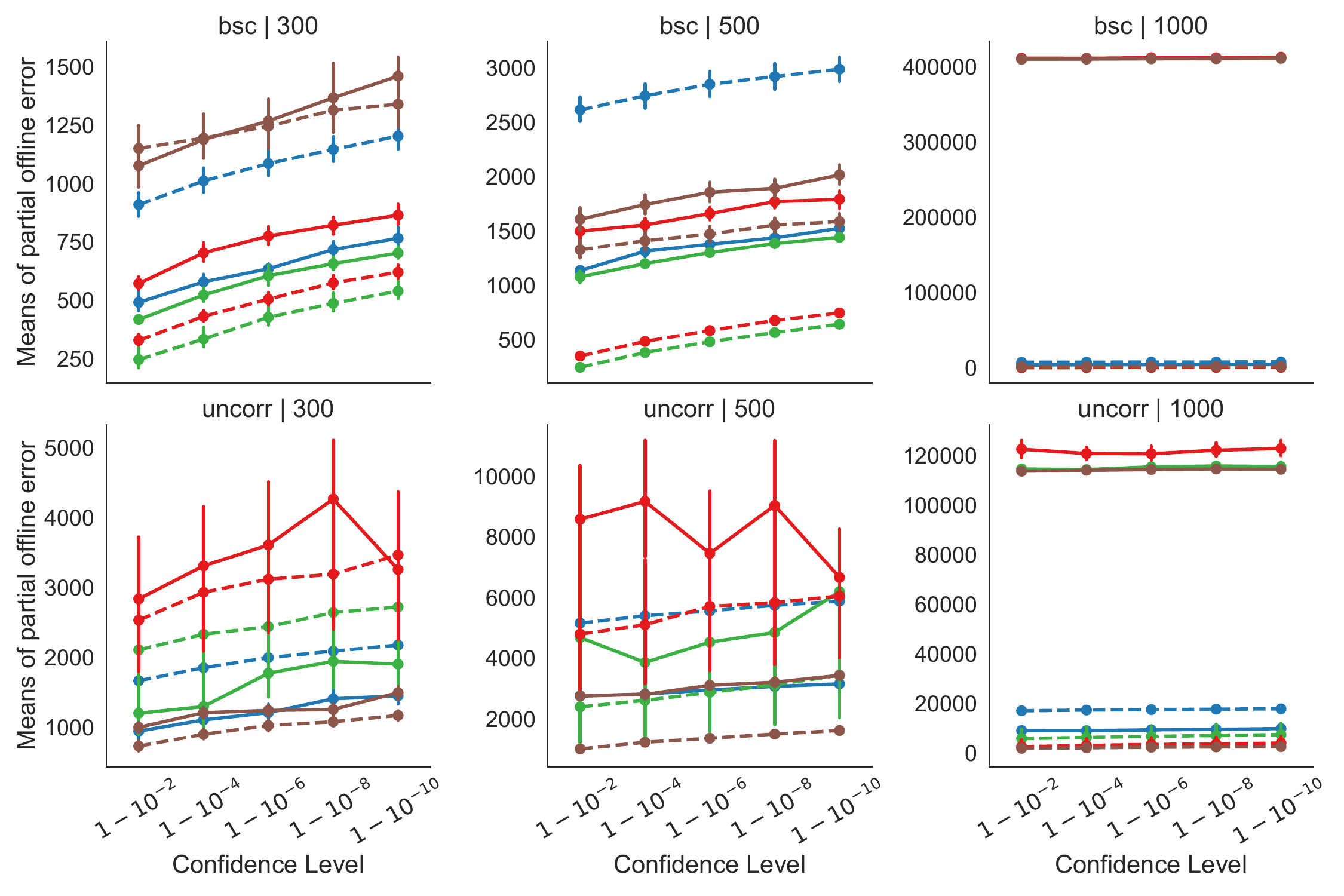}
        \caption{$\nu=10$ and the variance of each item is chosen according to $V_1$.}
        \label{fig: r_2000_t_10_var1}
    \end{subfigure}
    \begin{subfigure}[b]{0.48\textwidth}
        \centering
        \includegraphics[width=\textwidth]{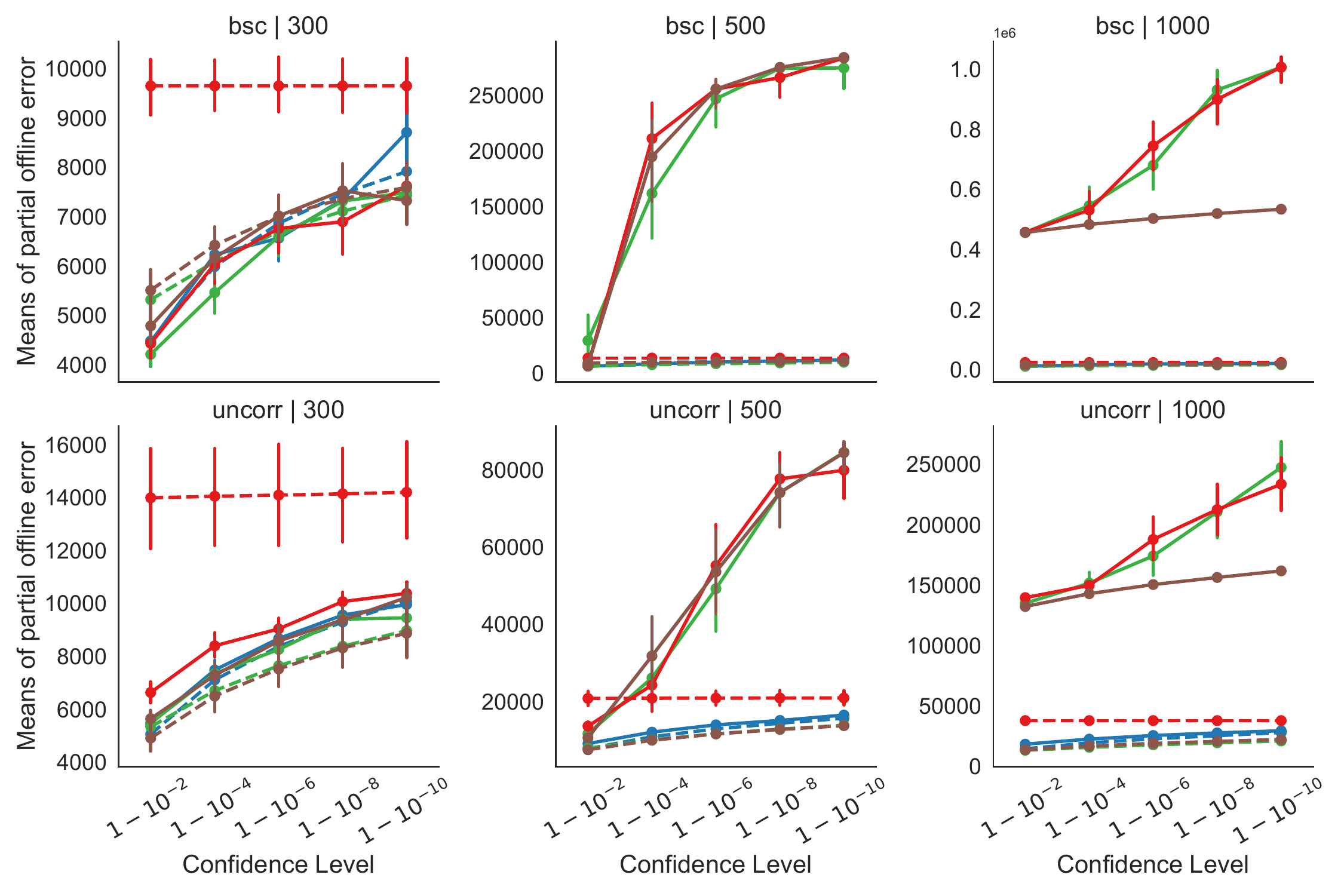}
        \caption{$\nu=10$ and the variance of each item is chosen according to $V_2$.}
        \label{fig: r_2000_t_10_var2}
    \end{subfigure}
    
    \begin{subfigure}[b]{0.48\textwidth}
        \centering
        \includegraphics[width=\textwidth]{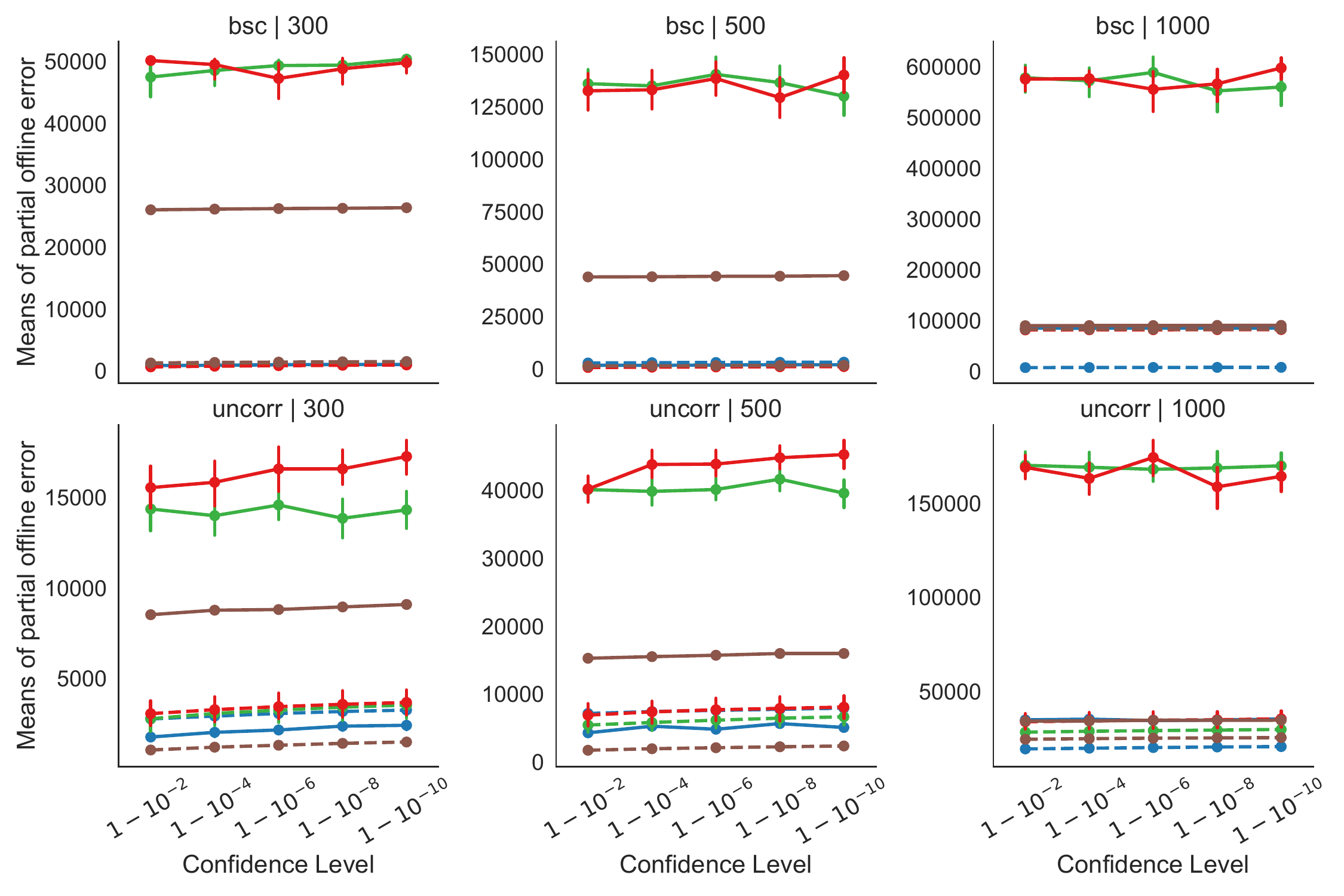}
        \caption{$\nu=50$ and the variance of each item is chosen according to $V_1$.}
        \label{fig: r_2000_t_50_var1}
    \end{subfigure}
    \begin{subfigure}[b]{0.48\textwidth}
        \centering
        \includegraphics[width=\textwidth]{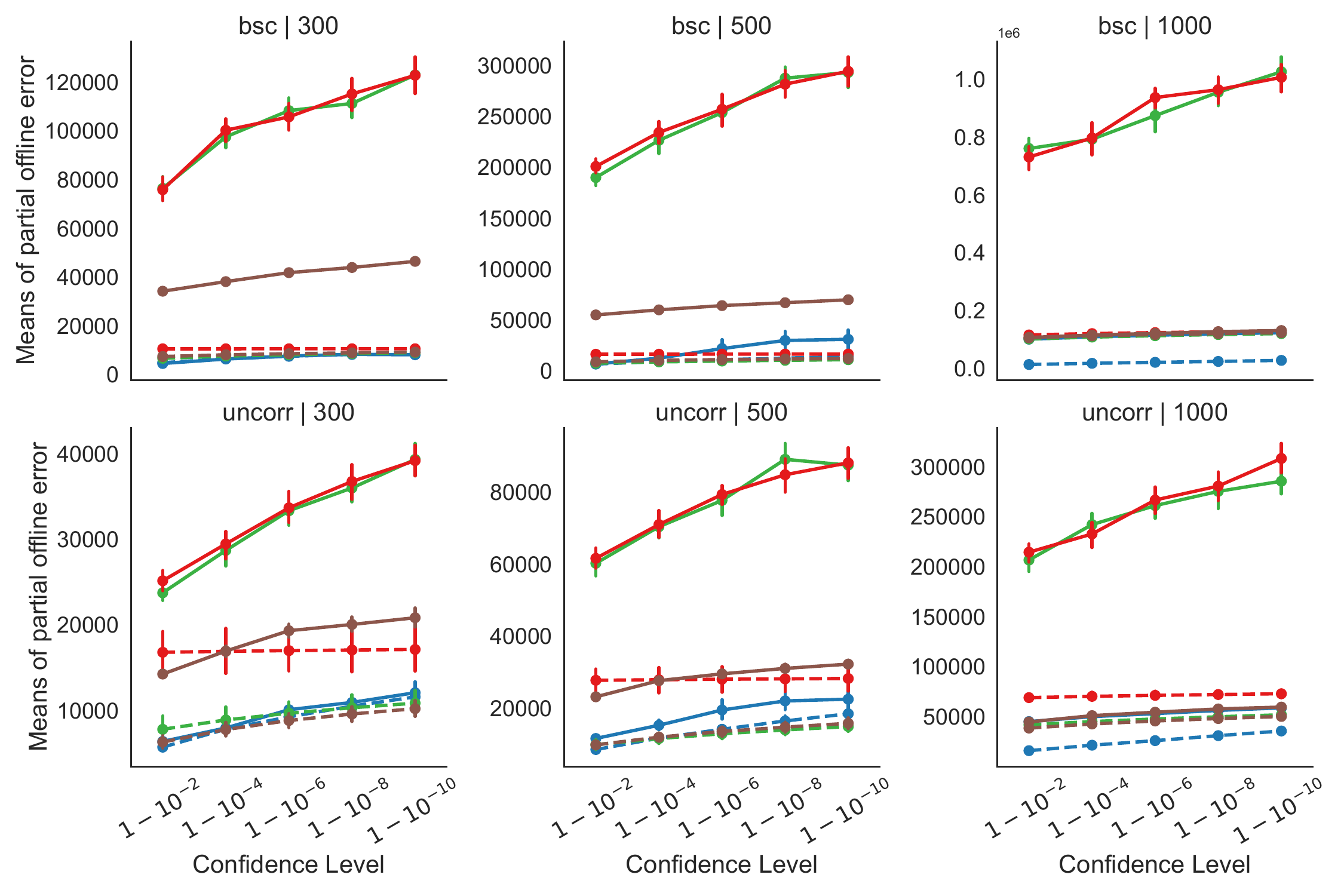}
        \caption{$\nu=50$ and the variance of each item is chosen according to $V_2$.}
        \label{fig: r_2000_t_50_var2}
    \end{subfigure}

    \begin{subfigure}[b]{0.48\textwidth}
        \centering
        \includegraphics[width=\textwidth]{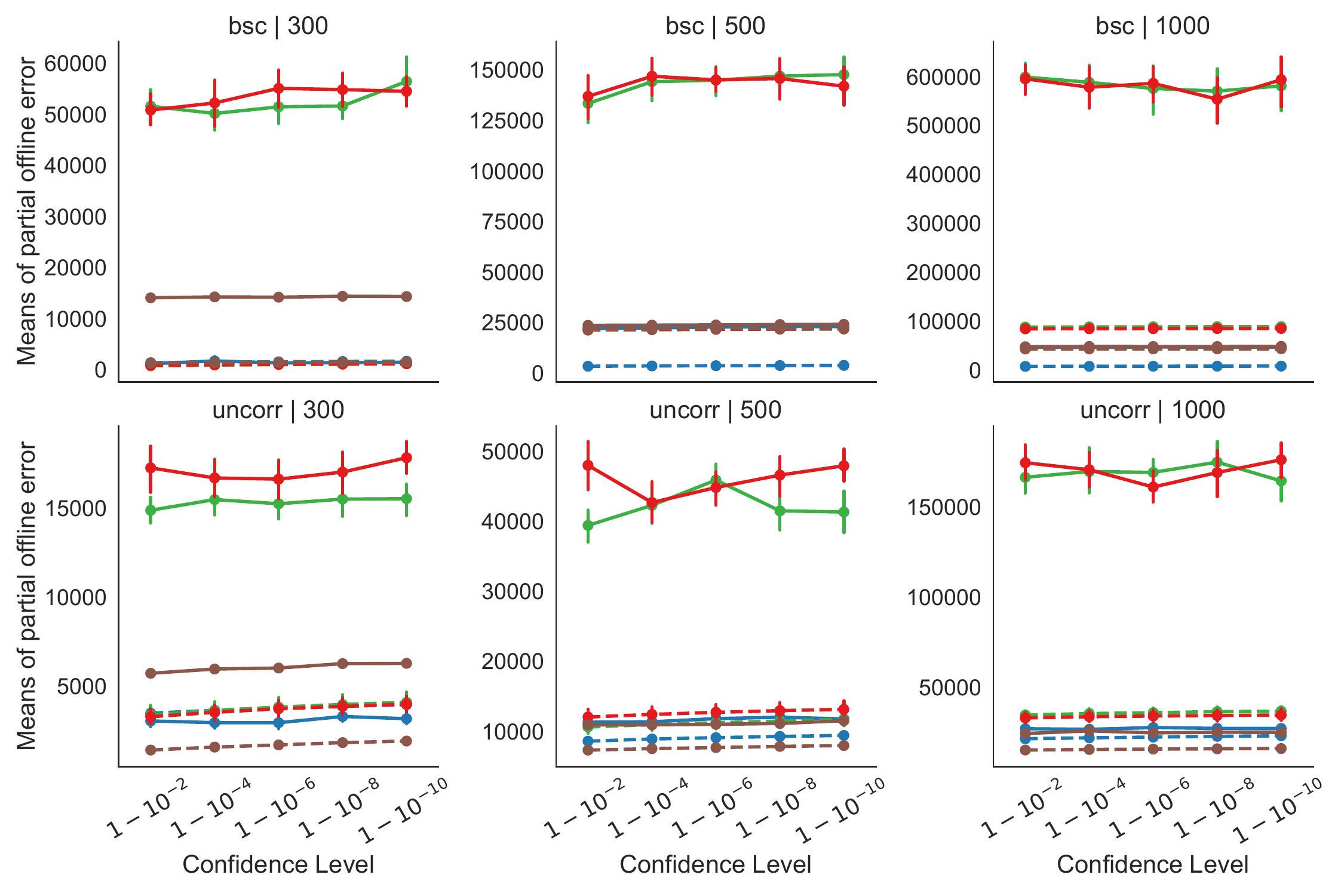}
        \caption{$\nu=100$ and the variance of each item is chosen according to $V_1$.}
        \label{fig: r_2000_t_100_var1}
    \end{subfigure}
    \begin{subfigure}[b]{0.48\textwidth}
        \centering
        \includegraphics[width=\textwidth]{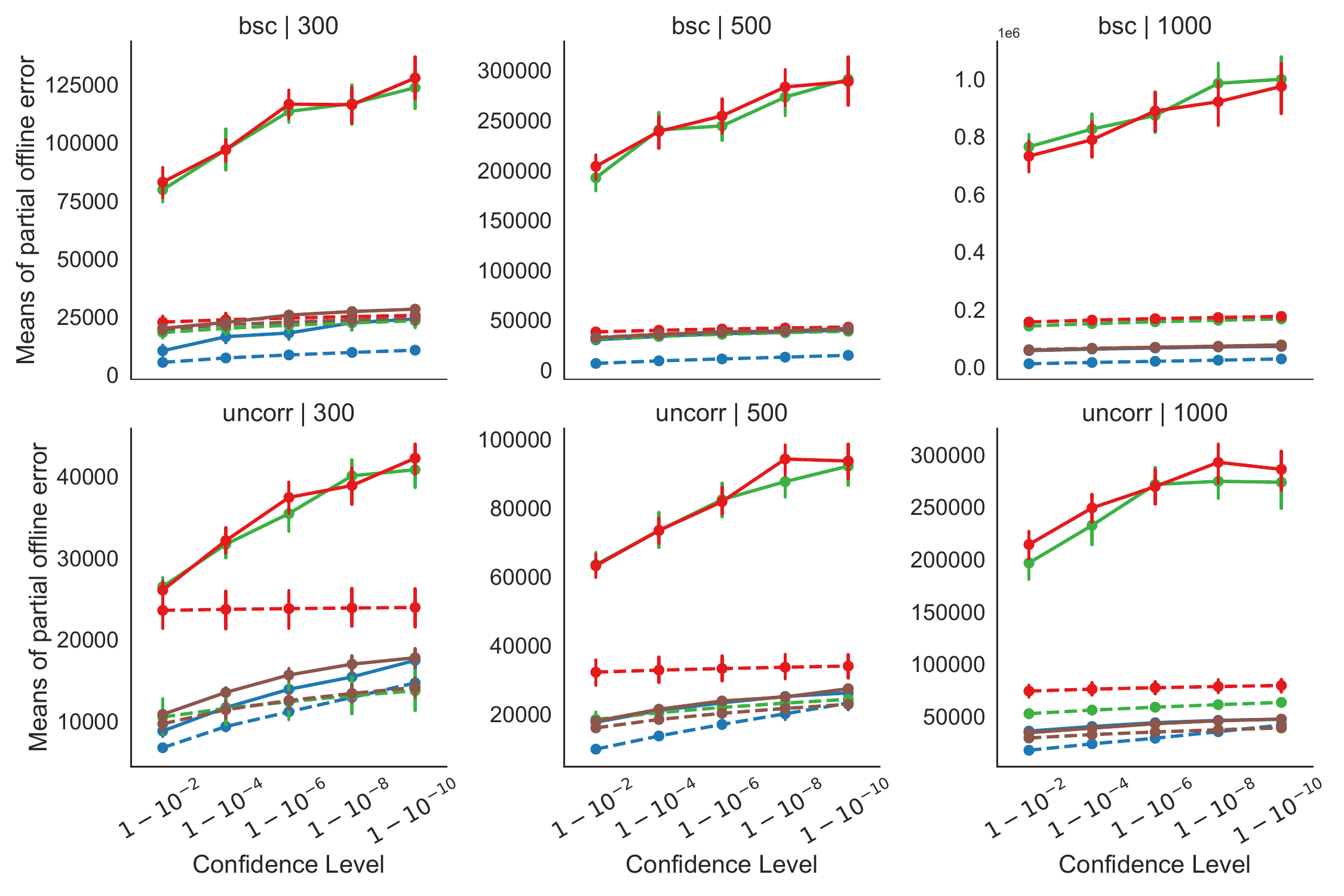}
        \caption{$\nu=100$ and the variance of each item is chosen according to $V_1$.}
        \label{fig: r_2000_t_100_var2}
    \end{subfigure}
    \begin{subfigure}[b]{\textwidth}
        \centering
        \includegraphics[scale=0.4]{ledgend.png}
    \end{subfigure}
    \caption{The mean value and the standard deviation of the total offline error for different algorithms when the magnitude of dynamic knapsack bound change follows $\mathcal{U}({-2000}, 2000)$}
    \label{fig: r_2000}
\end{figure*}

Figures \ref{fig: r_500} and \ref{fig: r_2000} summarize our findings. These figures present the mean value and the standard deviation of the partial offline error for 30 independent runs for different algorithms when the magnitude of dynamic change, $r$, is set to $500$ and $2000$ respectively. We use the same instance for all runs in the same setting, but the changes of the dynamic bound are independent in each run. The analysis considers the settings where the frequency of dynamic change, $\nu$, is set to $10$ (first row), $50$ (second row), $100$ (third row), and the variance of each item, $\sigma^2$, is chosen from either $V_1$ (left column) or $V_2$ (right column). Each sub-figure illustrates the results for bsc~(first row) and uncorr~(second row) type instances with $300$,$500$, and $1000$ items on average partial offline error (y-axis) with respective to confidence level $\alpha$. The lower offline error is, the better the performance is, as it indicates that the algorithm achieved a result closer to the optimal value $P(x^*)$.

The results with a small magnitude of the dynamic constraint changes are shown in Figure \ref{fig: r_500}. As shown in Figure~\ref{fig: r_500_t_10_var1}, when the variance of each item aligns with setting $V_1$ and frequency of dynamic changes is small~(i.e. $\nu = 10$), the MOEA/D\textsubscript{te} with a 3-objective formulation outperforms other algorithms for all the problems in both types. Figure~\ref{fig: r_500_t_10_var2} illustrates that the 3-objective approach performs better on all instances except the $300$ bsc instances, where the confidence level is looser with $\nu = 10$ and stochastic setting $V_2$. It is shown in Figure~\ref{fig: r_500_t_50_var1} that the 3-objective formulation with the MOEA/D\textsubscript{te} has better results with $300$ and $500$ instances while the GSEMO is superior when we have $1000$ items  for both bsc and uncorr problems. Figure~\ref{fig: r_500_t_50_var2} presents that the 3-objective formulation demonstrates the best performance in all settings when $\nu = 50$ and the variance is set according to $V_2$. When the frequency of dynamic changes is high~(i.e. $\nu=100$), the 3-objective formulation outperforms 2-objective formulation across both variance settings $V_1$~(Figures~\ref{fig: r_500_t_100_var1}) and $V_2$~(Figures~\ref{fig: r_500_t_100_var2}) for all problem instances.

Figure \ref{fig: r_2000} illustrates the mean value and the standard deviation of the offline error for different algorithms when the magnitude of the dynamic knapsack bound change follows $\mathcal{U}({-2000}, 2000)$. When the variance of each item is chosen according the the setting $V_1$ and with $\nu = 10$, for bsc and uncorr type problems with $300$, $500$, and $1000$ items, the MOEA/D\textsubscript{te} and the MOEA/D\textsubscript{ws} with the 3-objective formulation outperforms other algorithms respectively~(Figure~\ref{fig: r_2000_t_10_var1}). For $\nu = 50$ under $V_1$, the MOEA/D\textsubscript{pbi} and the MOEA/D\textsubscript{ws} with a 3-objective formulations yields the lowest offline errors at all confidence levels for $300$ and $500$ bsc and uncorr instances respectively, while the GSEMO with a 3-objective formulation is superior for the $1000$ instances of both types as illustrate in Figure~\ref{fig: r_2000_t_50_var1}. It is shown in Figure~\ref{fig: r_2000_t_100_var1} that with frequent changes (i.e., $\nu = 100$), the MOEA/D\textsubscript{te} and the GSEMO, both with a 3-objective formulation, excel for 300 and 500, 1000 item bsc instances, respectively. The MOEA/D\textsubscript{ws} with a 3-objective formulation stands out for all the uncorr instances. When the variance of each item aligns with the setting $V_2$, a 3-objective formulation surpasses other approaches in all benchmark instances and stochastic settings, except for the bsc 300 item instances when the number of dynamic changes are small and medium~(i.e. $\nu$ = 10, 50) as shown in Figures~\ref{fig: r_2000_t_10_var2}, \ref{fig: r_2000_t_50_var2} and~\ref{fig: r_2000_t_100_var2}.

These results reflect that when the variance of each item increases, the profit of the best feasible solution decreases for the same confidence level $\alpha$ as the uncertainty of the problem becoming large. Therefore, the average of partial offline error increases with the setting $V_2$ compared to $V_1$. It is shown that when the chance constrained confidence level becomes tighter, the partial offline error increases. This makes sense as tighten constraint on $\alpha$ allows the algorithms to compute solutions that are far from the bound $B$. Therefore the best obtained profit is decreased which tends to increase the offline error. As the number of dynamic changes, denoted by $\nu$, increases, we observe an increase in partial offline error. This trend occurs because more frequent dynamic changes provide less time for the algorithms to evolve and to find feasible solutions. MOEA/D\textsubscript{te} and MOEA/D\textsubscript{pbi} with a 2-objective formulation exhibit larger partial offline errors in most cases. The reason for this is these algorithms were not able to find feasible solutions before the occurrence of a dynamic change. Our results indicate that the proposed 3-objective formulation significantly outperforms the 2-objective formulation with more frequent dynamic changes with both stochastic settings. When the number of dynamic changes are small and medium, the 3-objective formulation has significant advantage except in some scenarios where the variance of each item is large and the number of items of the problem is small. Furthermore, the 3-objective formulation is highly effective when using this formulation to obtain a set of solutions that contains an optimal solution for any possible confidence level imposed on the constraint. Also, if we want to compute the optimal solution for a new confidence level within the interval, we do not need to compute again in the 3-objective model as we obtain a set of non-dominated solutions for any possible confidence level at once. Overall, we can state that the dynamic 3-objective formulation has promising results over 2-objective formulation to cater the dynamic and stochastic components effectively. 

\section{Conclusion}
\label{Sec: Conclusion}

In this paper, we evaluated the performance of 3-objective evolutionary approaches for solving the dynamic chance constrained knapsack problem (DCCKP), characterized by normally distributed random variables and knapsack bound changes over time. This problem formulation has wide applicability in various real-world scenarios as we can compute  a set of solutions that contains an optimal solution for any possible confidence level imposed on the constraint at once. We also conducted a comparative analysis with the 2-objective formulation which is limited to single confidence level $\alpha$. We evaluate these formulations on two MOEAs, namely the GSEMO and the three MOEA/D variants under small, medium and large frequency of knapsack bound change, small and large variances on items, different types and instances of problems. Our findings demonstrate that the 3-objective formulation significantly outperforms the 2-objective counterpart. Also, it is highly effective to compute an optimal solution for any possible confidence level imposed on the constraint at once. This result emphasizes the efficacy of the 3-objective approach in addressing complex, real-world optimization problems under dynamic and stochastic conditions. 

\section*{Acknowledgements}
This work has been supported by the Australian Research Council through grants DP190103894 and FT200100536.

\bibliographystyle{unsrtnat}
\bibliography{main} 

\appendix
\section{Appendix}

Corresponding to Figures~\ref{fig: r_500} and \ref{fig: r_2000} in the paper, we present more details on the experimental results in Tables \ref{table:2a_b} - \ref{table:3f_u}. Each column in the tables describes below. $file-n$ is the instance type(bsc and uncorr) and the number of items of each instance($300$, $500$ and $1000$). $\alpha$ is the confidence level imposed on the constraint and we present the results for the confidence level, $\alpha \in  \{1-10^{-2},1-10^{-4},1-10^{-6},1-10^{-8},1-10^{-10}\}$. stat column represents the measurement of the evaluations of the results we used, mean, standard deviation and the statistical comparison. Each column represent the algorithms we used namely, GSEMO, and MOEA/D with three different decomposition approaches, Tchebycheff ($MOEA/D_{te}$), PBI ($MOEA/D_{pbi}$), and weighted sum ($MOEA/D_{ws}$). $2D$ and $3D$ represents the results 2-and 3-objective formulations for each algorithm respectively. 

\begin{landscape}
\begin{table*}
\centering
\caption{Table corresponding to the results Figure 2(a) with bsc type instances, with setting $r= 500$ , $\nu = 10$, $\sigma_i^2 \in V_1$. The mean, standard deviation values and statistical tests of the partial offline error for GSEMO and MOEA/D variants for 2- and 3- formulations.}
\label{table:2a_b}
\begin{adjustbox}{max width=1.35\textheight}
\begin{tabular}{r|r|r|rr|rr|rr|rr}
\hline
\multirow{3}{*}{\rotatebox[origin=c]{90}{file-n}} & \multirow{3}{*}{$\alpha$} &\multirow{3}{*}{$stats$} & \multicolumn{2}{c}{$GSEMO$} & \multicolumn{2}{c}{$MOEA/D_{te}$} & \multicolumn{2}{c}{$MOEA/D_{pbi}$} & \multicolumn{2}{c}{$MOEA/D_{ws}$}\\
& & & $2D\textsuperscript{(1)}$ &  $3D\textsuperscript{(2)}$ & $2D\textsuperscript{(3)}$ & $3D\textsuperscript{(4)}$& $2D\textsuperscript{(5)}$ &$3D\textsuperscript{(6)}$ & $2D\textsuperscript{(7)}$ & $3D\textsuperscript{(8)}$ \\
\hline
\multirow{15}{*}{\rotatebox[origin=c]{90}{bsc-300}}  & 1-1e-2 & mean &\textbf{316.24} & 597.35 & 612.67 & \cellcolor{gray!25}\textbf{193.60} & 904.55 & \textbf{215.11} & \textbf{675.87} & 721.64\\
 & & sd &38.77 & 84.69 & 99.08 & 10.64 & 140.03 & 13.07 & 106.45 & 92.76\\
 & & stat &2\textsuperscript{(-)}3\textsuperscript{(-)}4\textsuperscript{(+)}5\textsuperscript{(-)} 7\textsuperscript{(-)}8\textsuperscript{(-)} & 1\textsuperscript{(+)} 4\textsuperscript{(+)}5\textsuperscript{(-)}6\textsuperscript{(+)}   & 1\textsuperscript{(+)} 4\textsuperscript{(+)}5\textsuperscript{(-)}6\textsuperscript{(+)}   & 1\textsuperscript{(-)}2\textsuperscript{(-)}3\textsuperscript{(-)}5\textsuperscript{(-)} 7\textsuperscript{(-)}8\textsuperscript{(-)} & 1\textsuperscript{(+)}2\textsuperscript{(+)}3\textsuperscript{(+)}4\textsuperscript{(+)}6\textsuperscript{(+)}   &  2\textsuperscript{(-)}3\textsuperscript{(-)} 5\textsuperscript{(-)}7\textsuperscript{(-)}8\textsuperscript{(-)} & 1\textsuperscript{(+)}  4\textsuperscript{(+)} 6\textsuperscript{(+)}  & 1\textsuperscript{(+)}  4\textsuperscript{(+)} 6\textsuperscript{(+)} \\
 \cline{2-11}
  & 1-1e-4 & mean &\textbf{437.90} & 709.31 & 748.46 & \cellcolor{gray!25}\textbf{302.30} & 1023.02 & \textbf{319.86} & \textbf{752.58} & 771.72\\
 & & sd &74.30 & 88.64 & 95.74 & 13.03 & 119.88 & 18.38 & 98.15 & 85.64\\
\hline
\end{tabular}
\end{adjustbox}
\end{table*}

\begin{table*}[!hbt]
\centering
\caption{Table corresponding to the results Figure 2(a) with uncorr type instances, with setting $r= 500$ , $\nu = 10$, $\sigma_i^2 \in V_1$. 
The mean, standard deviation values and statistical tests of the partial offline error for GSEMO and MOEA/D variants for 2- and 3- formulations. 
}
\label{table:2a_u}
\begin{adjustbox}{max width=1.35\textheight}
\begin{tabular}{r|r|r|rr|rr|rr|rr}
\hline
\multirow{3}{*}{\rotatebox[origin=c]{90}{file-n}} &   \multirow{3}{*}{$\alpha$} &\multirow{3}{*}{$stats$} & \multicolumn{2}{c}{$GSEMO$} & \multicolumn{2}{c}{$MOEA/D_{te}$} & \multicolumn{2}{c}{$MOEA/D_{pbi}$} & \multicolumn{2}{c}{$MOEA/D_{ws}$}\\
& & & $2D\textsuperscript{(1)}$ &  $3D\textsuperscript{(2)}$ & $2D\textsuperscript{(3)}$ & $3D\textsuperscript{(4)}$& $2D\textsuperscript{(5)}$ &$3D\textsuperscript{(6)}$ & $2D\textsuperscript{(7)}$ & $3D\textsuperscript{(8)}$ \\
\hline
\multirow{15}{*}{\rotatebox[origin=c]{90}{uncorr-300}}  & 1-1e-2 & mean &\textbf{431.00} & 919.12 & 765.05 & \cellcolor{gray!25}\textbf{297.27} & 1262.00 & \textbf{329.14} & 1029.29 & \textbf{696.00}\\
 & & sd &36.10 & 129.48 & 89.28 & 18.60 & 387.61 & 27.94 & 200.92 & 199.76\\
 & & stat &2\textsuperscript{(-)}3\textsuperscript{(-)} 5\textsuperscript{(-)} 7\textsuperscript{(-)}  & 1\textsuperscript{(+)} 4\textsuperscript{(+)} 6\textsuperscript{(+)}   & 1\textsuperscript{(+)} 4\textsuperscript{(+)}5\textsuperscript{(-)}6\textsuperscript{(+)}   &  2\textsuperscript{(-)}3\textsuperscript{(-)}5\textsuperscript{(-)} 7\textsuperscript{(-)}8\textsuperscript{(-)} & 1\textsuperscript{(+)} 3\textsuperscript{(+)}4\textsuperscript{(+)}6\textsuperscript{(+)} 8\textsuperscript{(+)} &  2\textsuperscript{(-)}3\textsuperscript{(-)} 5\textsuperscript{(-)}7\textsuperscript{(-)}8\textsuperscript{(-)} & 1\textsuperscript{(+)}  4\textsuperscript{(+)} 6\textsuperscript{(+)}8\textsuperscript{(+)} &    4\textsuperscript{(+)}5\textsuperscript{(-)}6\textsuperscript{(+)}7\textsuperscript{(-)}\\
 \cline{2-11}
  & 1-1e-4 & mean &\textbf{618.71} & 1093.34 & 929.44 & \cellcolor{gray!25}\textbf{479.52} & 1173.46 & \textbf{507.00} & 1179.65 & \textbf{818.30}\\
 & & sd &42.79 & 136.15 & 101.98 & 30.93 & 137.67 & 35.21 & 190.85 & 198.36\\
\hline
\end{tabular}
\end{adjustbox}
\end{table*}

\begin{table*}[!hbt]
\centering
\caption{Table corresponding to the results Figure 2(b) with bsc type instances, with setting $r= 500$ , $\nu = 10$, $\sigma_i^2 \in V_2$. The mean, standard deviation values and statistical tests of the partial offline error for GSEMO and MOEA/D variants for 2- and 3- formulations. 
}
\label{table:2b_b}
\begin{adjustbox}{max width=1.35\textheight}
\begin{tabular}{r|r|r|rr|rr|rr|rr}
\hline
\multirow{3}{*}{\rotatebox[origin=c]{90}{file-n}} &   \multirow{3}{*}{$\alpha$} &\multirow{3}{*}{$stats$} & \multicolumn{2}{c}{$GSEMO$} & \multicolumn{2}{c}{$MOEA/D_{te}$} & \multicolumn{2}{c}{$MOEA/D_{pbi}$} & \multicolumn{2}{c}{$MOEA/D_{ws}$}\\
& & & $2D\textsuperscript{(1)}$ &  $3D\textsuperscript{(2)}$ & $2D\textsuperscript{(3)}$ & $3D\textsuperscript{(4)}$& $2D\textsuperscript{(5)}$ &$3D\textsuperscript{(6)}$ & $2D\textsuperscript{(7)}$ & $3D\textsuperscript{(8)}$ \\
\hline
\multirow{15}{*}{\rotatebox[origin=c]{90}{bsc-300}} & 1-1e-2 & mean &4873.54 & \textbf{4706.82} & \textbf{4555.65} & 5004.89 & \textbf{4830.48} & 8789.68 & \cellcolor{gray!25}\textbf{4495.56} & 5308.15\\
 & & sd &312.79 & 176.18 & 195.10 & 386.77 & 235.74 & 567.13 & 205.74 & 245.06\\
 & & stat & 3\textsuperscript{(+)}  6\textsuperscript{(-)}7\textsuperscript{(+)}8\textsuperscript{(-)} &     6\textsuperscript{(-)} 8\textsuperscript{(-)} & 1\textsuperscript{(-)} 4\textsuperscript{(-)}5\textsuperscript{(-)}6\textsuperscript{(-)} 8\textsuperscript{(-)} &   3\textsuperscript{(+)} 6\textsuperscript{(-)}7\textsuperscript{(+)}  &   3\textsuperscript{(+)} 6\textsuperscript{(-)}7\textsuperscript{(+)}8\textsuperscript{(-)} & 1\textsuperscript{(+)}2\textsuperscript{(+)}3\textsuperscript{(+)}4\textsuperscript{(+)}5\textsuperscript{(+)}7\textsuperscript{(+)}  & 1\textsuperscript{(-)}  4\textsuperscript{(-)}5\textsuperscript{(-)}6\textsuperscript{(-)}8\textsuperscript{(-)} & 1\textsuperscript{(+)}2\textsuperscript{(+)}3\textsuperscript{(+)} 5\textsuperscript{(+)} 7\textsuperscript{(+)}\\
 \cline{2-11} 
 & 1-1e-4 & mean &6603.34 & \textbf{6297.61} & 5995.28 & \textbf{5859.96} & \textbf{6218.14} & 8789.68 & \cellcolor{gray!25}\textbf{5833.22} & 6763.95\\
 & & sd &372.01 & 251.15 & 268.66 & 340.11 & 289.61 & 567.13 & 276.58 & 391.51\\
\hline
\end{tabular}
\end{adjustbox}
\end{table*}

\begin{table*}[!hbt]
\centering
\caption{Table corresponding to the results Figure 2(b) with uncorr type instances, with setting $r= 500$ , $\nu = 10$, $\sigma_i^2 \in V_2$. The mean, standard deviation values and statistical tests of the partial offline error for GSEMO and MOEA/D variants for 2- and 3- formulations. 
}
\label{table:2b_u}
\begin{adjustbox}{max width=1.35\textheight}
\begin{tabular}{r|r|r|rr|rr|rr|rr}
\hline
\multirow{3}{*}{\rotatebox[origin=c]{90}{file-n}} &   \multirow{3}{*}{$\alpha$} &\multirow{3}{*}{$stats$} & \multicolumn{2}{c}{$GSEMO$} & \multicolumn{2}{c}{$MOEA/D_{te}$} & \multicolumn{2}{c}{$MOEA/D_{pbi}$} & \multicolumn{2}{c}{$MOEA/D_{ws}$}\\
& & & $2D\textsuperscript{(1)}$ &  $3D\textsuperscript{(2)}$ & $2D\textsuperscript{(3)}$ & $3D\textsuperscript{(4)}$& $2D\textsuperscript{(5)}$ &$3D\textsuperscript{(6)}$ & $2D\textsuperscript{(7)}$ & $3D\textsuperscript{(8)}$ \\
\hline
\multirow{15}{*}{\rotatebox[origin=c]{90}{uncorr-300}} & 1-1e-2 & mean &5378.31 & \textbf{5253.42} & 5218.89 & \cellcolor{gray!25}\textbf{5033.43} & \textbf{5544.74} & 12079.86 & 5464.15 & \textbf{5213.17}\\
 & & sd &229.76 & 81.18 & 183.15 & 234.74 & 237.74 & 1755.28 & 176.96 & 194.72\\
 & & stat &  4\textsuperscript{(+)} 6\textsuperscript{(-)}   &    5\textsuperscript{(-)}6\textsuperscript{(-)}7\textsuperscript{(-)}  &    5\textsuperscript{(-)}6\textsuperscript{(-)}7\textsuperscript{(-)}  & 1\textsuperscript{(-)}  5\textsuperscript{(-)}6\textsuperscript{(-)}7\textsuperscript{(-)}  &  2\textsuperscript{(+)}3\textsuperscript{(+)}4\textsuperscript{(+)}6\textsuperscript{(-)} 8\textsuperscript{(+)} & 1\textsuperscript{(+)}2\textsuperscript{(+)}3\textsuperscript{(+)}4\textsuperscript{(+)}5\textsuperscript{(+)}7\textsuperscript{(+)}8\textsuperscript{(+)} &  2\textsuperscript{(+)}3\textsuperscript{(+)}4\textsuperscript{(+)} 6\textsuperscript{(-)}8\textsuperscript{(+)} &     5\textsuperscript{(-)}6\textsuperscript{(-)}7\textsuperscript{(-)}\\
 \cline{2-11} 
 & 1-1e-4 &  mean &7801.37 & \textbf{7506.02} & 7268.21 & \cellcolor{gray!25}\textbf{6866.66} & \textbf{7608.41} & 12079.86 & 7622.53 & \textbf{7189.44}\\
 & & sd &394.61 & 113.61 & 204.73 & 171.74 & 301.42 & 1755.28 & 285.96 & 269.46\\
\hline
\end{tabular}
\end{adjustbox}
\end{table*}

\begin{table*}[!hbt]
\centering
\caption{Table corresponding to the results Figure 2(c) with bsc type instances, with setting $r= 500$ , $\nu = 50$, $\sigma_i^2 \in V_1$. The mean, standard deviation values and statistical tests of the partial offline error for GSEMO and MOEA/D variants for 2- and 3- formulations. 
}
\label{table:2c_b}
\begin{adjustbox}{max width=1.35\textheight}
\begin{tabular}{r|r|r|rr|rr|rr|rr}
\hline
\multirow{3}{*}{\rotatebox[origin=c]{90}{file-n}} &\multirow{3}{*}{$\alpha$} &\multirow{3}{*}{$stats$} & \multicolumn{2}{c}{$GSEMO$} & \multicolumn{2}{c}{$MOEA/D_{te}$} & \multicolumn{2}{c}{$MOEA/D_{pbi}$} & \multicolumn{2}{c}{$MOEA/D_{ws}$}\\
& & & $2D\textsuperscript{(1)}$ &  $3D\textsuperscript{(2)}$ & $2D\textsuperscript{(3)}$ & $3D\textsuperscript{(4)}$& $2D\textsuperscript{(5)}$ &$3D\textsuperscript{(6)}$ & $2D\textsuperscript{(7)}$ & $3D\textsuperscript{(8)}$ \\
\hline
\multirow{15}{*}{\rotatebox[origin=c]{90}{bsc-300}} & 1-1e-2 &  mean &\textbf{471.56} & 669.23 & 49882.01 & \cellcolor{gray!25}\textbf{260.07} & 48775.29 & \textbf{333.68} & 25679.81 & \textbf{749.70}\\
& & sd &138.52 & 95.08 & 178.39 & 9.65 & 6152.27 & 64.78 & 100.34 & 95.79\\
  & & stat & 3\textsuperscript{(-)}4\textsuperscript{(+)}5\textsuperscript{(-)} 7\textsuperscript{(-)}  &  3\textsuperscript{(-)}4\textsuperscript{(+)}5\textsuperscript{(-)}6\textsuperscript{(+)}   & 1\textsuperscript{(+)}2\textsuperscript{(+)}4\textsuperscript{(+)} 6\textsuperscript{(+)} 8\textsuperscript{(+)} & 1\textsuperscript{(-)}2\textsuperscript{(-)}3\textsuperscript{(-)}5\textsuperscript{(-)} 7\textsuperscript{(-)}8\textsuperscript{(-)} & 1\textsuperscript{(+)}2\textsuperscript{(+)} 4\textsuperscript{(+)}6\textsuperscript{(+)}7\textsuperscript{(+)}8\textsuperscript{(+)} &  2\textsuperscript{(-)}3\textsuperscript{(-)} 5\textsuperscript{(-)}7\textsuperscript{(-)}8\textsuperscript{(-)} & 1\textsuperscript{(+)}  4\textsuperscript{(+)}5\textsuperscript{(-)}6\textsuperscript{(+)}  &   3\textsuperscript{(-)}4\textsuperscript{(+)}5\textsuperscript{(-)}6\textsuperscript{(+)} \\
 \cline{2-11}
  & 1-1e-4 & mean &\textbf{561.47} & 779.34 & 49279.87 & \cellcolor{gray!25}\textbf{373.66} & 49709.52 & \textbf{447.93} & 25793.21 & \textbf{826.53}\\
& & sd &44.77 & 99.40 & 4445.57 & 12.51 & 4343.65 & 63.59 & 89.38 & 89.85\\
\hline
\end{tabular}
\end{adjustbox}
\end{table*}

\begin{table*}[!hbt]
\centering
\caption{Table corresponding to the results Figure 2(c) with uncorr type instances, with setting $r= 500$ , $\nu = 50$, $\sigma_i^2 \in V_1$. The mean, standard deviation values and statistical tests of the partial offline error for GSEMO and MOEA/D variants for 2- and 3- formulations. 
}
\label{table:2c_u}
\begin{adjustbox}{max width=1.35\textheight}
\begin{tabular}{r|r|r|rr|rr|rr|rr}
\hline
\multirow{3}{*}{\rotatebox[origin=c]{90}{file-n}} &\multirow{3}{*}{$\alpha$} &\multirow{3}{*}{$stats$} & \multicolumn{2}{c}{$GSEMO$} & \multicolumn{2}{c}{$MOEA/D_{te}$} & \multicolumn{2}{c}{$MOEA/D_{pbi}$} & \multicolumn{2}{c}{$MOEA/D_{ws}$}\\
& & & $2D\textsuperscript{(1)}$ &  $3D\textsuperscript{(2)}$ & $2D\textsuperscript{(3)}$ & $3D\textsuperscript{(4)}$& $2D\textsuperscript{(5)}$ &$3D\textsuperscript{(6)}$ & $2D\textsuperscript{(7)}$ & $3D\textsuperscript{(8)}$ \\
\hline
\multirow{15}{*}{\rotatebox[origin=c]{90}{uncorr-300}} & 1-1e-2 & mean &\textbf{747.78} & 1131.17 & 14316.99 & \cellcolor{gray!25}\textbf{550.00} & 15358.32 & \textbf{1460.62} & 8184.91 & \textbf{804.09}\\
& & sd &101.38 & 119.98 & 1231.78 & 466.73 & 2005.89 & 2318.35 & 189.47 & 126.23\\
  & & stat & 3\textsuperscript{(-)} 5\textsuperscript{(-)} 7\textsuperscript{(-)}  &  3\textsuperscript{(-)}4\textsuperscript{(+)}5\textsuperscript{(-)}6\textsuperscript{(-)}   & 1\textsuperscript{(+)}2\textsuperscript{(+)}4\textsuperscript{(+)} 6\textsuperscript{(+)} 8\textsuperscript{(+)} &  2\textsuperscript{(-)}3\textsuperscript{(-)}5\textsuperscript{(-)} 7\textsuperscript{(-)}  & 1\textsuperscript{(+)}2\textsuperscript{(+)} 4\textsuperscript{(+)}6\textsuperscript{(+)} 8\textsuperscript{(+)} &  2\textsuperscript{(+)}3\textsuperscript{(-)} 5\textsuperscript{(-)}7\textsuperscript{(-)}  & 1\textsuperscript{(+)}  4\textsuperscript{(+)} 6\textsuperscript{(+)}8\textsuperscript{(+)} &   3\textsuperscript{(-)} 5\textsuperscript{(-)} 7\textsuperscript{(-)}\\
 \cline{2-11}
 & 1-1e-4 & mean &\textbf{971.79} & 1305.80 & 14193.68 & \cellcolor{gray!25}\textbf{790.40} & 15720.24 & \textbf{1757.84} & 8324.45 & \textbf{973.24}\\
& & sd &132.24 & 117.03 & 1610.42 & 596.56 & 1366.34 & 2443.66 & 122.71 & 121.14\\
\hline
\end{tabular}
\end{adjustbox}
\end{table*}
\begin{table*}[!hbt]
\centering
\caption{Table corresponding to the results Figure 2(d) with bsc type instances, with setting $r= 500$ , $\nu = 50$, $\sigma_i^2 \in V_2$. The mean, standard deviation values and statistical tests of the partial offline error for GSEMO and MOEA/D variants for 2- and 3- formulations. 
}

\label{table:2d_b}
\begin{adjustbox}{max width=1.35\textheight}
\begin{tabular}{r|r|r|rr|rr|rr|rr}
\hline
\multirow{3}{*}{\rotatebox[origin=c]{90}{file-n}} & \multirow{3}{*}{$\alpha$} &\multirow{3}{*}{$stats$} & \multicolumn{2}{c}{$GSEMO$} & \multicolumn{2}{c}{$MOEA/D_{te}$} & \multicolumn{2}{c}{$MOEA/D_{pbi}$} & \multicolumn{2}{c}{$MOEA/D_{ws}$}\\
& & & $2D\textsuperscript{(1)}$ &  $3D\textsuperscript{(2)}$ & $2D\textsuperscript{(3)}$ & $3D\textsuperscript{(4)}$& $2D\textsuperscript{(5)}$ &$3D\textsuperscript{(6)}$ & $2D\textsuperscript{(7)}$ & $3D\textsuperscript{(8)}$ \\
\hline
\multirow{15}{*}{\rotatebox[origin=c]{90}{bsc-300}} &  1-1e-2 & mean &5153.53 & \cellcolor{gray!25}\textbf{4847.37} & 84486.71 & \textbf{5727.18} & 85760.33 & \textbf{10213.13} & 34471.94 & \textbf{5981.36}\\
 & & sd &455.72 & 628.07 & 12153.53 & 648.17 & 11410.73 & 1268.16 & 473.22 & 861.48\\
 & & stat & 3\textsuperscript{(-)} 5\textsuperscript{(-)}6\textsuperscript{(-)}7\textsuperscript{(-)}  &  3\textsuperscript{(-)} 5\textsuperscript{(-)}6\textsuperscript{(-)}7\textsuperscript{(-)}  & 1\textsuperscript{(+)}2\textsuperscript{(+)}4\textsuperscript{(+)} 6\textsuperscript{(+)} 8\textsuperscript{(+)} &   3\textsuperscript{(-)}5\textsuperscript{(-)}6\textsuperscript{(-)}7\textsuperscript{(-)}  & 1\textsuperscript{(+)}2\textsuperscript{(+)} 4\textsuperscript{(+)}6\textsuperscript{(+)} 8\textsuperscript{(+)} & 1\textsuperscript{(+)}2\textsuperscript{(+)}3\textsuperscript{(-)}4\textsuperscript{(+)}5\textsuperscript{(-)}   & 1\textsuperscript{(+)}2\textsuperscript{(+)} 4\textsuperscript{(+)}  8\textsuperscript{(+)} &   3\textsuperscript{(-)} 5\textsuperscript{(-)} 7\textsuperscript{(-)}\\
 \cline{2-11}
 & 1-1e-4 & mean &9910.64 & \cellcolor{gray!25}\textbf{6440.24} & 95311.36 & \textbf{6613.88} & 94437.52 & \textbf{10213.13} & 38738.81 & \textbf{7644.74}\\
 & & sd &12336.03 & 861.48 & 11744.14 & 611.99 & 12562.29 & 1268.16 & 690.46 & 963.62\\
\hline
\end{tabular}
\end{adjustbox}
\end{table*}

\begin{table*}[!hbt]
\centering
\caption{Table corresponding to the results Figure 2(d) with uncorr type instances, with setting $r= 500$ , $\nu = 50$, $\sigma_i^2 \in V_2$. The mean, standard deviation values and statistical tests of the partial offline error for GSEMO and MOEA/D variants for 2- and 3- formulations. 
}
\label{table:2d_u}
\begin{adjustbox}{max width=1.35\textheight}
\begin{tabular}{r|r|r|rr|rr|rr|rr}
\hline
\multirow{3}{*}{\rotatebox[origin=c]{90}{file-n}} & \multirow{3}{*}{$\alpha$} &\multirow{3}{*}{$stats$} & \multicolumn{2}{c}{$GSEMO$} & \multicolumn{2}{c}{$MOEA/D_{te}$} & \multicolumn{2}{c}{$MOEA/D_{pbi}$} & \multicolumn{2}{c}{$MOEA/D_{ws}$}\\
& & & $2D\textsuperscript{(1)}$ &  $3D\textsuperscript{(2)}$ & $2D\textsuperscript{(3)}$ & $3D\textsuperscript{(4)}$& $2D\textsuperscript{(5)}$ &$3D\textsuperscript{(6)}$ & $2D\textsuperscript{(7)}$ & $3D\textsuperscript{(8)}$ \\
\hline

\multirow{24}{*}{\rotatebox[origin=c]{90}{uncorr-300}} &  1-1e-2 & mean &5870.15 & \cellcolor{gray!25}\textbf{5389.01} & 23698.75 & \textbf{6281.42} & 25995.30 & \textbf{16210.31} & 14121.28 & \textbf{5800.91}\\
 & & sd &515.81 & 541.56 & 3254.97 & 956.70 & 3474.38 & 2722.10 & 428.56 & 537.48\\
 & & stat & 3\textsuperscript{(-)} 5\textsuperscript{(-)}6\textsuperscript{(-)}7\textsuperscript{(-)}  &  3\textsuperscript{(-)} 5\textsuperscript{(-)}6\textsuperscript{(-)}7\textsuperscript{(-)}  & 1\textsuperscript{(+)}2\textsuperscript{(+)}4\textsuperscript{(+)}  7\textsuperscript{(+)}8\textsuperscript{(+)} &   3\textsuperscript{(-)}5\textsuperscript{(-)}6\textsuperscript{(-)}7\textsuperscript{(-)}  & 1\textsuperscript{(+)}2\textsuperscript{(+)} 4\textsuperscript{(+)}6\textsuperscript{(+)}7\textsuperscript{(+)}8\textsuperscript{(+)} & 1\textsuperscript{(+)}2\textsuperscript{(+)} 4\textsuperscript{(+)}5\textsuperscript{(-)} 8\textsuperscript{(+)} & 1\textsuperscript{(+)}2\textsuperscript{(+)}3\textsuperscript{(-)}4\textsuperscript{(+)}5\textsuperscript{(-)} 8\textsuperscript{(+)} &   3\textsuperscript{(-)} 5\textsuperscript{(-)}6\textsuperscript{(-)}7\textsuperscript{(-)}\\
 \cline{2-11}
 & 1-1e-4 & mean &8698.97 & \cellcolor{gray!25}\textbf{7492.13} & 30150.37 & \textbf{7777.96} & 32716.05 & \textbf{16216.82} & 16963.81 & \textbf{7791.65}\\
 & & sd &2277.13 & 744.39 & 3668.25 & 846.74 & 2352.15 & 2707.83 & 507.44 & 740.86\\
\hline
\end{tabular}
\end{adjustbox}
\end{table*}
\begin{table*}[!hbt]
\centering
\caption{Table corresponding to the results Figure 2(e) with bsc type instances, with setting $r= 500$ , $\nu = 100$, $\sigma_i^2 \in V_1$. The mean, standard deviation values and statistical tests of the partial offline error for GSEMO and MOEA/D variants for 2- and 3- formulations. 
}

\label{table:2e_b}
\begin{adjustbox}{max width=1.35\textheight}
\begin{tabular}{r|r|r|rr|rr|rr|rr}
\hline
\multirow{3}{*}{\rotatebox[origin=c]{90}{file-n}} & \multirow{3}{*}{$\alpha$} &\multirow{3}{*}{$stats$} & \multicolumn{2}{c}{$GSEMO$} & \multicolumn{2}{c}{$MOEA/D_{te}$} & \multicolumn{2}{c}{$MOEA/D_{pbi}$} & \multicolumn{2}{c}{$MOEA/D_{ws}$}\\
& & & $2D\textsuperscript{(1)}$ &  $3D\textsuperscript{(2)}$ & $2D\textsuperscript{(3)}$ & $3D\textsuperscript{(4)}$& $2D\textsuperscript{(5)}$ &$3D\textsuperscript{(6)}$ & $2D\textsuperscript{(7)}$ & $3D\textsuperscript{(8)}$ \\
\hline
\multirow{15}{*}{\rotatebox[origin=c]{90}{bsc-300}} &  1-1e-2 & mean &\textbf{521.63} & 666.10 & 54841.80 & \textbf{713.58} & 56997.88 & \cellcolor{gray!25}\textbf{441.28} & 13622.97 & \textbf{1151.55}\\
 & & sd &82.13 & 74.62 & 8695.60 & 2200.49 & 8891.06 & 119.10 & 72.58 & 2194.43\\
 & & stat & 3\textsuperscript{(-)} 5\textsuperscript{(-)} 7\textsuperscript{(-)}8\textsuperscript{(-)} &  3\textsuperscript{(-)}4\textsuperscript{(-)}5\textsuperscript{(-)} 7\textsuperscript{(-)}  & 1\textsuperscript{(+)}2\textsuperscript{(+)}4\textsuperscript{(+)} 6\textsuperscript{(+)} 8\textsuperscript{(+)} &  2\textsuperscript{(+)}3\textsuperscript{(-)}5\textsuperscript{(-)} 7\textsuperscript{(-)}8\textsuperscript{(-)} & 1\textsuperscript{(+)}2\textsuperscript{(+)} 4\textsuperscript{(+)}6\textsuperscript{(+)} 8\textsuperscript{(+)} &   3\textsuperscript{(-)} 5\textsuperscript{(-)}7\textsuperscript{(-)}8\textsuperscript{(-)} & 1\textsuperscript{(+)}2\textsuperscript{(+)} 4\textsuperscript{(+)} 6\textsuperscript{(+)}  & 1\textsuperscript{(+)} 3\textsuperscript{(-)}4\textsuperscript{(+)}5\textsuperscript{(-)}6\textsuperscript{(+)} \\
 \cline{2-11}
 & 1-1e-4 & mean &1018.74 & \textbf{776.10} & 54558.35 & \textbf{830.57} & 54613.57 & \cellcolor{gray!25}\textbf{557.82} & 13727.36 & \textbf{1234.96}\\
 & & sd &2213.73 & 81.57 & 8703.21 & 2204.23 & 10634.43 & 117.70 & 67.07 & 2198.27\\
\hline
\end{tabular}
\end{adjustbox}
\end{table*}

\begin{table*}[!hbt]
\centering
\caption{Table corresponding to the results Figure 2(e) with uncorr type instances, with setting $r= 500$ , $\nu = 100$, $\sigma_i^2 \in V_1$. The mean, standard deviation values and statistical tests of the partial offline error for GSEMO and MOEA/D variants for 2- and 3- formulations. 
}
\label{table:2e_u}
\begin{adjustbox}{max width=1.35\textheight}
\begin{tabular}{r|r|r|rr|rr|rr|rr}
\hline
\multirow{3}{*}{\rotatebox[origin=c]{90}{file-n}} & \multirow{3}{*}{$\alpha$} &\multirow{3}{*}{$stats$} & \multicolumn{2}{c}{$GSEMO$} & \multicolumn{2}{c}{$MOEA/D_{te}$} & \multicolumn{2}{c}{$MOEA/D_{pbi}$} & \multicolumn{2}{c}{$MOEA/D_{ws}$}\\
& & & $2D\textsuperscript{(1)}$ &  $3D\textsuperscript{(2)}$ & $2D\textsuperscript{(3)}$ & $3D\textsuperscript{(4)}$& $2D\textsuperscript{(5)}$ &$3D\textsuperscript{(6)}$ & $2D\textsuperscript{(7)}$ & $3D\textsuperscript{(8)}$ \\
\hline
\multirow{24}{*}{\rotatebox[origin=c]{90}{uncorr-300}} &  1-1e-2 & mean &\textbf{1194.96} & 1298.69 & 14731.77 & \textbf{1291.65} & 16530.64 & \textbf{1601.41} & 5165.77 & \cellcolor{gray!25}\textbf{931.75}\\
 & & sd &829.44 & 135.35 & 2074.54 & 1037.81 & 3298.54 & 1975.07 & 216.34 & 77.99\\
 & & stat & 3\textsuperscript{(-)} 5\textsuperscript{(-)} 7\textsuperscript{(-)}  &  3\textsuperscript{(-)} 5\textsuperscript{(-)}    & 1\textsuperscript{(+)}2\textsuperscript{(+)}4\textsuperscript{(+)} 6\textsuperscript{(+)} 8\textsuperscript{(+)} &   3\textsuperscript{(-)}5\textsuperscript{(-)} 7\textsuperscript{(-)}  & 1\textsuperscript{(+)}2\textsuperscript{(+)} 4\textsuperscript{(+)}6\textsuperscript{(+)} 8\textsuperscript{(+)} &   3\textsuperscript{(-)} 5\textsuperscript{(-)}7\textsuperscript{(-)}  & 1\textsuperscript{(+)}  4\textsuperscript{(+)} 6\textsuperscript{(+)}8\textsuperscript{(+)} &   3\textsuperscript{(-)} 5\textsuperscript{(-)} 7\textsuperscript{(-)}\\
 \cline{2-11}
 & 1-1e-4 & mean &\textbf{1277.26} & 1477.28 & 15532.48 & \textbf{1577.80} & 16734.71 & \textbf{1828.75} & 5390.36 & \cellcolor{gray!25}\textbf{1090.70}\\
 & & sd &588.56 & 134.15 & 1781.25 & 1092.25 & 2645.36 & 2031.92 & 244.02 & 83.84\\
\hline
\end{tabular}
\end{adjustbox}
\end{table*}
\begin{table*}[!hbt]
\centering
\caption{Table corresponding to the results Figure 2(f) with bsc type instances, with setting $r= 500$ , $\nu = 100$, $\sigma_i^2 \in V_2$. The mean, standard deviation values and statistical tests of the partial offline error for GSEMO and MOEA/D variants for 2- and 3- formulations. 
}
\label{table:2f_b}
\begin{adjustbox}{max width=1.35\textheight}
\begin{tabular}{r|r|r|rr|rr|rr|rr}
\hline
\multirow{3}{*}{\rotatebox[origin=c]{90}{file-n}} & \multirow{3}{*}{$\alpha$} &\multirow{3}{*}{$stats$} & \multicolumn{2}{c}{$GSEMO$} & \multicolumn{2}{c}{$MOEA/D_{te}$} & \multicolumn{2}{c}{$MOEA/D_{pbi}$} & \multicolumn{2}{c}{$MOEA/D_{ws}$}\\
& & & $2D\textsuperscript{(1)}$ &  $3D\textsuperscript{(2)}$ & $2D\textsuperscript{(3)}$ & $3D\textsuperscript{(4)}$& $2D\textsuperscript{(5)}$ &$3D\textsuperscript{(6)}$ & $2D\textsuperscript{(7)}$ & $3D\textsuperscript{(8)}$ \\
\hline
\multirow{15}{*}{\rotatebox[origin=c]{90}{bsc-300}} &  1-1e-2 & mean &11912.63 & \cellcolor{gray!25}\textbf{5320.42} & 80504.45 & \textbf{20390.60} & 82676.45 & \textbf{24335.96} & 20592.37 & \textbf{20410.26}\\
 & & sd &7261.24 & 776.67 & 15969.60 & 1520.60 & 16926.96 & 3590.40 & 623.11 & 998.10\\
 & & stat & 3\textsuperscript{(-)}4\textsuperscript{(-)}5\textsuperscript{(-)}6\textsuperscript{(-)}7\textsuperscript{(-)}  &  3\textsuperscript{(-)}4\textsuperscript{(-)}5\textsuperscript{(-)}6\textsuperscript{(-)}7\textsuperscript{(-)}8\textsuperscript{(-)} & 1\textsuperscript{(+)}2\textsuperscript{(+)}4\textsuperscript{(+)}  7\textsuperscript{(+)}8\textsuperscript{(+)} & 1\textsuperscript{(+)}2\textsuperscript{(+)}3\textsuperscript{(-)}5\textsuperscript{(-)}    & 1\textsuperscript{(+)}2\textsuperscript{(+)} 4\textsuperscript{(+)} 7\textsuperscript{(+)}8\textsuperscript{(+)} & 1\textsuperscript{(+)}2\textsuperscript{(+)}    8\textsuperscript{(+)} & 1\textsuperscript{(+)}2\textsuperscript{(+)}3\textsuperscript{(-)} 5\textsuperscript{(-)}   &  2\textsuperscript{(+)}3\textsuperscript{(-)} 5\textsuperscript{(-)}6\textsuperscript{(-)} \\
 \cline{2-11} 
  & 1-1e-4 & mean &22667.16 & \cellcolor{gray!25}\textbf{7043.80} & 104442.72 & \textbf{22483.41} & 101681.54 & \textbf{25495.62} & 23680.44 & \textbf{23156.07}\\
 & & sd &928.43 & 1058.50 & 11775.67 & 1555.20 & 18170.88 & 3869.03 & 981.19 & 1136.12\\
\hline
\end{tabular}
\end{adjustbox}
\end{table*}

\begin{table*}[!hbt]
\centering
\caption{Table corresponding to the results Figure 2(f) with uncorr type instances, with setting $r= 500$ , $\nu = 100$, $\sigma_i^2 \in V_2$. The mean, standard deviation values and statistical tests of the partial offline error for GSEMO and MOEA/D variants for 2- and 3- formulations. 
}
\label{table:2f_u}
\begin{adjustbox}{max width=1.35\textheight}
\begin{tabular}{r|r|r|rr|rr|rr|rr}
\hline
\multirow{3}{*}{\rotatebox[origin=c]{90}{file-n}} & \multirow{3}{*}{$\alpha$} &\multirow{3}{*}{$stats$} & \multicolumn{2}{c}{$GSEMO$} & \multicolumn{2}{c}{$MOEA/D_{te}$} & \multicolumn{2}{c}{$MOEA/D_{pbi}$} & \multicolumn{2}{c}{$MOEA/D_{ws}$}\\
& & & $2D\textsuperscript{(1)}$ &  $3D\textsuperscript{(2)}$ & $2D\textsuperscript{(3)}$ & $3D\textsuperscript{(4)}$& $2D\textsuperscript{(5)}$ &$3D\textsuperscript{(6)}$ & $2D\textsuperscript{(7)}$ & $3D\textsuperscript{(8)}$ \\
\hline
\multirow{15}{*}{\rotatebox[origin=c]{90}{uncorr-300}} &  1-1e-2 & mean &8430.68 & \cellcolor{gray!25}\textbf{5447.73} & 26352.35 & \textbf{9738.50} & 26693.53 & \textbf{20910.20} & 10724.70 & \textbf{9392.54}\\
 & & sd &1879.33 & 609.50 & 3791.44 & 2630.33 & 3780.11 & 3714.26 & 532.37 & 1657.24\\
 & & stat & 3\textsuperscript{(-)} 5\textsuperscript{(-)}6\textsuperscript{(-)}   &  3\textsuperscript{(-)}4\textsuperscript{(-)}5\textsuperscript{(-)}6\textsuperscript{(-)}7\textsuperscript{(-)}8\textsuperscript{(-)} & 1\textsuperscript{(+)}2\textsuperscript{(+)}4\textsuperscript{(+)}  7\textsuperscript{(+)}8\textsuperscript{(+)} &  2\textsuperscript{(+)}3\textsuperscript{(-)}5\textsuperscript{(-)}6\textsuperscript{(-)}   & 1\textsuperscript{(+)}2\textsuperscript{(+)} 4\textsuperscript{(+)} 7\textsuperscript{(+)}8\textsuperscript{(+)} & 1\textsuperscript{(+)}2\textsuperscript{(+)} 4\textsuperscript{(+)}  8\textsuperscript{(+)} &  2\textsuperscript{(+)}3\textsuperscript{(-)} 5\textsuperscript{(-)}   &  2\textsuperscript{(+)}3\textsuperscript{(-)} 5\textsuperscript{(-)}6\textsuperscript{(-)} \\
 \cline{2-11} 
  & 1-1e-4 & mean &11594.38 & \cellcolor{gray!25}\textbf{7507.36} & 32700.50 & \textbf{11276.97} & 33314.83 & \textbf{21177.84} & 13277.92 & \textbf{11680.45}\\
 & & sd &2160.03 & 855.84 & 4382.14 & 2520.67 & 4232.50 & 3715.91 & 877.94 & 1785.90\\
\hline
\end{tabular}
\end{adjustbox}
\end{table*}

\begin{table*}[!hbt]
\centering
\caption{Table corresponding to the results Figure 3(a) with bsc type instances, with setting $r= 2000$ , $\nu = 10$, $\sigma_i^2 \in V_1$. The mean, standard deviation values and statistical tests of the partial offline error for GSEMO and MOEA/D variants for 2- and 3- formulations. 
}

\label{table:3a_b}
\begin{adjustbox}{max width=1.35\textheight}
\begin{tabular}{r|r|r|rr|rr|rr|rr}
\hline
\multirow{3}{*}{\rotatebox[origin=c]{90}{file-n}} & \multirow{3}{*}{$\alpha$} &\multirow{3}{*}{$stats$} & \multicolumn{2}{c}{$GSEMO$} & \multicolumn{2}{c}{$MOEA/D_{te}$} & \multicolumn{2}{c}{$MOEA/D_{pbi}$} & \multicolumn{2}{c}{$MOEA/D_{ws}$}\\
& & & $2D\textsuperscript{(1)}$ &  $3D\textsuperscript{(2)}$ & $2D\textsuperscript{(3)}$ & $3D\textsuperscript{(4)}$& $2D\textsuperscript{(5)}$ &$3D\textsuperscript{(6)}$ & $2D\textsuperscript{(7)}$ & $3D\textsuperscript{(8)}$ \\
\hline
\multirow{15}{*}{\rotatebox[origin=c]{90}{bsc-300}} & 1-1e-2 & mean &\textbf{491.83} & 910.47 & 418.46 & \cellcolor{gray!25}\textbf{246.37} & 573.13 & \textbf{329.25} & \textbf{1077.41} & 1151.65\\
  & & sd &125.04 & 143.31 & 40.53 & 128.00 & 79.11 & 64.80 & 243.76 & 294.18\\
 & & stat &2\textsuperscript{(-)} 4\textsuperscript{(+)}  7\textsuperscript{(-)}8\textsuperscript{(-)} & 1\textsuperscript{(+)}3\textsuperscript{(+)}4\textsuperscript{(+)} 6\textsuperscript{(+)}   &  2\textsuperscript{(-)}   7\textsuperscript{(-)}8\textsuperscript{(-)} & 1\textsuperscript{(-)}2\textsuperscript{(-)} 5\textsuperscript{(-)} 7\textsuperscript{(-)}8\textsuperscript{(-)} &    4\textsuperscript{(+)}6\textsuperscript{(+)}7\textsuperscript{(-)}8\textsuperscript{(-)} &  2\textsuperscript{(-)}  5\textsuperscript{(-)}7\textsuperscript{(-)}8\textsuperscript{(-)} & 1\textsuperscript{(+)} 3\textsuperscript{(+)}4\textsuperscript{(+)}5\textsuperscript{(+)}6\textsuperscript{(+)}  & 1\textsuperscript{(+)} 3\textsuperscript{(+)}4\textsuperscript{(+)}5\textsuperscript{(+)}6\textsuperscript{(+)} \\
 \cline{2-11} 
  & 1-1e-4 & mean &\textbf{580.00} & 1012.48 & 523.24 & \cellcolor{gray!25}\textbf{334.67} & 703.74 & \textbf{431.92} & \textbf{1189.61} & 1196.12\\
  & & sd &84.78 & 148.91 & 82.33 & 120.99 & 114.06 & 65.80 & 245.92 & 275.28\\
\hline
\end{tabular}
\end{adjustbox}
\end{table*}

\begin{table*}[!hbt]
\centering
\caption{Table corresponding to the results Figure 3(a) with uncorr type instances, with setting $r= 2000$ , $\nu = 10$, $\sigma_i^2 \in V_1$. The mean, standard deviation values and statistical tests of the partial offline error for GSEMO and MOEA/D variants for 2- and 3- formulations. 
}
\label{table:3a_u}
\begin{adjustbox}{max width=1.33\textheight}
\begin{tabular}{r|r|r|rr|rr|rr|rr}
\hline
\multirow{3}{*}{\rotatebox[origin=c]{90}{file-n}} & \multirow{3}{*}{$\alpha$} &\multirow{3}{*}{$stats$} & \multicolumn{2}{c}{$GSEMO$} & \multicolumn{2}{c}{$MOEA/D_{te}$} & \multicolumn{2}{c}{$MOEA/D_{pbi}$} & \multicolumn{2}{c}{$MOEA/D_{ws}$}\\
& & & $2D\textsuperscript{(1)}$ &  $3D\textsuperscript{(2)}$ & $2D\textsuperscript{(3)}$ & $3D\textsuperscript{(4)}$& $2D\textsuperscript{(5)}$ &$3D\textsuperscript{(6)}$ & $2D\textsuperscript{(7)}$ & $3D\textsuperscript{(8)}$ \\
\hline
\multirow{15}{*}{\rotatebox[origin=c]{90}{uncorr-300}} &  1-1e-2 & mean &\textbf{945.98} & 1668.00 & \textbf{1202.33} & 2109.33 & 2838.76 & \textbf{2533.63} & 1000.74 & \cellcolor{gray!25}\textbf{731.26}\\
  & & sd &391.72 & 319.49 & 993.38 & 2035.29 & 2614.97 & 2211.14 & 212.12 & 205.92\\
 & & stat &2\textsuperscript{(-)}  5\textsuperscript{(-)}    & 1\textsuperscript{(+)}3\textsuperscript{(+)}    8\textsuperscript{(+)} &  2\textsuperscript{(-)} 5\textsuperscript{(-)}    &         & 1\textsuperscript{(+)} 3\textsuperscript{(+)}   8\textsuperscript{(+)} &       8\textsuperscript{(+)} &         &  2\textsuperscript{(-)}  5\textsuperscript{(-)}6\textsuperscript{(-)} \\
 \cline{2-11} 
  & 1-1e-4 & mean &\textbf{1106.87} & 1853.26 & \textbf{1297.71} & 2332.48 & 3311.04 & \textbf{2933.58} & 1209.59 & \cellcolor{gray!25}\textbf{901.43}\\
  & & sd &226.28 & 310.82 & 1074.81 & 1982.36 & 2476.86 & 2396.38 & 232.74 & 196.62\\
 & & stat &2\textsuperscript{(-)}  5\textsuperscript{(-)}    & 1\textsuperscript{(+)}3\textsuperscript{(+)}    8\textsuperscript{(+)} &  2\textsuperscript{(-)} 5\textsuperscript{(-)}6\textsuperscript{(-)}   &       8\textsuperscript{(+)} & 1\textsuperscript{(+)} 3\textsuperscript{(+)}   8\textsuperscript{(+)} &   3\textsuperscript{(+)}   8\textsuperscript{(+)} &         &  2\textsuperscript{(-)} 4\textsuperscript{(-)}5\textsuperscript{(-)}6\textsuperscript{(-)} \\
\hline
\end{tabular}
\end{adjustbox}
\end{table*}
\begin{table*}[!hbt]
\centering
\caption{Table corresponding to the results Figure 3(b) with bsc type instances, with setting $r= 2000$ , $\nu = 10$, $\sigma_i^2 \in V_2$. The mean, standard deviation values and statistical tests of the partial offline error for GSEMO and MOEA/D variants for 2- and 3- formulations. 
}

\label{table:3b_b}
\begin{adjustbox}{max width=1.35\textheight}
\begin{tabular}{r|r|r|rr|rr|rr|rr}
\hline
\multirow{3}{*}{\rotatebox[origin=c]{90}{file-n}} & \multirow{3}{*}{$\alpha$} &\multirow{3}{*}{$stats$} & \multicolumn{2}{c}{$GSEMO$} & \multicolumn{2}{c}{$MOEA/D_{te}$} & \multicolumn{2}{c}{$MOEA/D_{pbi}$} & \multicolumn{2}{c}{$MOEA/D_{ws}$}\\
& & & $2D\textsuperscript{(1)}$ &  $3D\textsuperscript{(2)}$ & $2D\textsuperscript{(3)}$ & $3D\textsuperscript{(4)}$& $2D\textsuperscript{(5)}$ &$3D\textsuperscript{(6)}$ & $2D\textsuperscript{(7)}$ & $3D\textsuperscript{(8)}$ \\
\hline
\multirow{15}{*}{\rotatebox[origin=c]{90}{bsc-300}} & 1-1e-2 & mean &\textbf{4466.61} & 4487.73 & \cellcolor{gray!25}\textbf{4215.97} & 5322.54 & \textbf{4437.00} & 9655.01 & \textbf{4791.60} & 5514.44\\
 & & sd &1014.97 & 957.71 & 664.10 & 1355.19 & 808.15 & 1545.97 & 950.87 & 1139.87\\
 & & stat &    6\textsuperscript{(-)}   &     6\textsuperscript{(-)} 8\textsuperscript{(-)} &   4\textsuperscript{(-)} 6\textsuperscript{(-)} 8\textsuperscript{(-)} &   3\textsuperscript{(+)} 6\textsuperscript{(-)}   &     6\textsuperscript{(-)} 8\textsuperscript{(-)} & 1\textsuperscript{(+)}2\textsuperscript{(+)}3\textsuperscript{(+)}4\textsuperscript{(+)}5\textsuperscript{(+)}7\textsuperscript{(+)}8\textsuperscript{(+)} &      6\textsuperscript{(-)}  &  2\textsuperscript{(+)}3\textsuperscript{(+)} 5\textsuperscript{(+)}6\textsuperscript{(-)} \\
 \cline{2-11} 
 & 1-1e-4 & mean &6237.01 & \textbf{5995.76} & \cellcolor{gray!25}\textbf{5466.07} & 6093.17 & \textbf{6045.73} & 9655.01 & \textbf{6168.59} & 6428.00\\
 & & sd &1174.91 & 1306.86 & 1123.67 & 1351.26 & 1030.07 & 1545.97 & 1154.45 & 1114.76\\
 & & stat &    6\textsuperscript{(-)}   &     6\textsuperscript{(-)}   &     6\textsuperscript{(-)}   &     6\textsuperscript{(-)}   &     6\textsuperscript{(-)}   & 1\textsuperscript{(+)}2\textsuperscript{(+)}3\textsuperscript{(+)}4\textsuperscript{(+)}5\textsuperscript{(+)}7\textsuperscript{(+)}8\textsuperscript{(+)} &      6\textsuperscript{(-)}  &      6\textsuperscript{(-)} \\
\hline
\end{tabular}
\end{adjustbox}
\end{table*}

\begin{table*}[!hbt]
\centering
\caption{Table corresponding to the results Figure 3(b) with uncorr type instances, with setting $r= 2000$ , $\nu = 10$, $\sigma_i^2 \in V_2$. The mean, standard deviation values and statistical tests of the partial offline error for GSEMO and MOEA/D variants for 2- and 3- formulations. 
}
\label{table:3b_u}
\begin{adjustbox}{max width=1.35\textheight}
\begin{tabular}{r|r|r|rr|rr|rr|rr}
\hline
\multirow{3}{*}{\rotatebox[origin=c]{90}{file-n}} & \multirow{3}{*}{$\alpha$} &\multirow{3}{*}{$stats$} & \multicolumn{2}{c}{$GSEMO$} & \multicolumn{2}{c}{$MOEA/D_{te}$} & \multicolumn{2}{c}{$MOEA/D_{pbi}$} & \multicolumn{2}{c}{$MOEA/D_{ws}$}\\
& & & $2D\textsuperscript{(1)}$ &  $3D\textsuperscript{(2)}$ & $2D\textsuperscript{(3)}$ & $3D\textsuperscript{(4)}$& $2D\textsuperscript{(5)}$ &$3D\textsuperscript{(6)}$ & $2D\textsuperscript{(7)}$ & $3D\textsuperscript{(8)}$ \\
\hline
\multirow{15}{*}{\rotatebox[origin=c]{90}{uncorr-300}} &  1-1e-2 & mean &5473.62 & \textbf{5066.08} & 5494.36 & \textbf{5347.63} & \textbf{6643.20} & 13997.21 & 5654.57 & \cellcolor{gray!25}\textbf{4922.01}\\
 & & sd &937.40 & 875.68 & 576.05 & 1647.00 & 1130.91 & 5332.64 & 993.99 & 1289.74\\
 & & stat &    6\textsuperscript{(-)}   &    5\textsuperscript{(-)}6\textsuperscript{(-)}   &    5\textsuperscript{(-)}6\textsuperscript{(-)}   &    5\textsuperscript{(-)}6\textsuperscript{(-)}   &  2\textsuperscript{(+)}3\textsuperscript{(+)}4\textsuperscript{(+)}  8\textsuperscript{(+)} & 1\textsuperscript{(+)}2\textsuperscript{(+)}3\textsuperscript{(+)}4\textsuperscript{(+)} 7\textsuperscript{(+)}8\textsuperscript{(+)} &      6\textsuperscript{(-)}  &     5\textsuperscript{(-)}6\textsuperscript{(-)} \\
 \cline{2-11} 
 & 1-1e-4 & mean &7491.01 & \textbf{7120.50} & 7356.04 & \textbf{6713.08} & \textbf{8406.00} & 14055.01 & 7294.12 & \cellcolor{gray!25}\textbf{6506.89}\\
 & & sd &1188.59 & 1299.54 & 854.95 & 1663.94 & 1307.81 & 5261.54 & 1390.90 & 1631.22\\
 & & stat &    6\textsuperscript{(-)}   &     6\textsuperscript{(-)}   &     6\textsuperscript{(-)}   &    5\textsuperscript{(-)}6\textsuperscript{(-)}   &    4\textsuperscript{(+)}  8\textsuperscript{(+)} & 1\textsuperscript{(+)}2\textsuperscript{(+)}3\textsuperscript{(+)}4\textsuperscript{(+)} 7\textsuperscript{(+)}8\textsuperscript{(+)} &      6\textsuperscript{(-)}  &     5\textsuperscript{(-)}6\textsuperscript{(-)} \\
\hline
\end{tabular}
\end{adjustbox}
\end{table*}
\begin{table*}[!hbt]
\centering
\caption{Table corresponding to the results Figure 3(c) with bsc type instances, with setting $r= 2000$ , $\nu = 50$, $\sigma_i^2 \in V_1$. The mean, standard deviation values and statistical tests of the partial offline error for GSEMO and MOEA/D variants for 2- and 3- formulations. 
}

\label{table:3c_b}
\begin{adjustbox}{max width=1.34\textheight}
\begin{tabular}{r|r|r|rr|rr|rr|rr}
\hline
\multirow{3}{*}{\rotatebox[origin=c]{90}{file-n}} & \multirow{3}{*}{$\alpha$} &\multirow{3}{*}{$stats$} & \multicolumn{2}{c}{$GSEMO$} & \multicolumn{2}{c}{$MOEA/D_{te}$} & \multicolumn{2}{c}{$MOEA/D_{pbi}$} & \multicolumn{2}{c}{$MOEA/D_{ws}$}\\
& & & $2D\textsuperscript{(1)}$ &  $3D\textsuperscript{(2)}$ & $2D\textsuperscript{(3)}$ & $3D\textsuperscript{(4)}$& $2D\textsuperscript{(5)}$ &$3D\textsuperscript{(6)}$ & $2D\textsuperscript{(7)}$ & $3D\textsuperscript{(8)}$ \\
\hline
\multirow{15}{*}{\rotatebox[origin=c]{90}{bsc-300}} & 1-1e-2 & mean &\textbf{915.70} & 1218.96 & 47504.92 & \textbf{739.76} & 50173.84 & \cellcolor{gray!25}\textbf{707.87} & 26064.57 & \textbf{1332.74}\\
 & sd &298.29 & 163.74 & 7397.13 & 513.76 & 344.26 & 313.88 & 247.48 & 338.84\\
 & stat & 3\textsuperscript{(-)} 5\textsuperscript{(-)} 7\textsuperscript{(-)}  &  3\textsuperscript{(-)} 5\textsuperscript{(-)} 7\textsuperscript{(-)}  & 1\textsuperscript{(+)}2\textsuperscript{(+)}4\textsuperscript{(+)} 6\textsuperscript{(+)} 8\textsuperscript{(+)} &   3\textsuperscript{(-)}5\textsuperscript{(-)} 7\textsuperscript{(-)}8\textsuperscript{(-)} & 1\textsuperscript{(+)}2\textsuperscript{(+)} 4\textsuperscript{(+)}6\textsuperscript{(+)} 8\textsuperscript{(+)} &   3\textsuperscript{(-)} 5\textsuperscript{(-)}7\textsuperscript{(-)}8\textsuperscript{(-)} & 1\textsuperscript{(+)}2\textsuperscript{(+)} 4\textsuperscript{(+)} 6\textsuperscript{(+)}8\textsuperscript{(+)} &   3\textsuperscript{(-)}4\textsuperscript{(+)}5\textsuperscript{(-)}6\textsuperscript{(+)}7\textsuperscript{(-)}\\
 \cline{2-11} 
 & 1-1e-4 & mean &\textbf{952.15} & 1327.12 & 48567.32 & \textbf{856.27} & 49513.23 & \cellcolor{gray!25}\textbf{814.84} & 26174.84 & \textbf{1414.52}\\
 & sd &199.79 & 177.48 & 6142.37 & 501.83 & 4408.48 & 318.92 & 218.50 & 356.32\\
\hline
\end{tabular}
\end{adjustbox}
\end{table*}

\begin{table*}[!hbt]
\centering
\caption{Table corresponding to the results Figure 3(c) with uncorr type instances, with setting $r= 2000$ , $\nu = 50$, $\sigma_i^2 \in V_1$. The mean, standard deviation values and statistical tests of the partial offline error for GSEMO and MOEA/D variants for 2- and 3- formulations. 
}
\label{table:3c_u}
\begin{adjustbox}{max width=1.35\textheight}
\begin{tabular}{r|r|r|rr|rr|rr|rr}
\hline
\multirow{3}{*}{\rotatebox[origin=c]{90}{file-n}} & \multirow{3}{*}{$\alpha$} &\multirow{3}{*}{$stats$} & \multicolumn{2}{c}{$GSEMO$} & \multicolumn{2}{c}{$MOEA/D_{te}$} & \multicolumn{2}{c}{$MOEA/D_{pbi}$} & \multicolumn{2}{c}{$MOEA/D_{ws}$}\\
& & & $2D\textsuperscript{(1)}$ &  $3D\textsuperscript{(2)}$ & $2D\textsuperscript{(3)}$ & $3D\textsuperscript{(4)}$& $2D\textsuperscript{(5)}$ &$3D\textsuperscript{(6)}$ & $2D\textsuperscript{(7)}$ & $3D\textsuperscript{(8)}$ \\
\hline
\multirow{15}{*}{\rotatebox[origin=c]{90}{uncorr-300}} &  1-1e-2 & mean &\textbf{1793.71} & 2792.87 & 14366.97 & \textbf{2801.39} & 15547.12 & \textbf{3081.34} & 8537.96 & \cellcolor{gray!25}\textbf{1078.84}\\
 & sd &632.59 & 423.82 & 3106.87 & 1833.90 & 3418.36 & 1994.75 & 227.04 & 171.37\\
 & stat & 3\textsuperscript{(-)} 5\textsuperscript{(-)} 7\textsuperscript{(-)}  &  3\textsuperscript{(-)} 5\textsuperscript{(-)} 7\textsuperscript{(-)}8\textsuperscript{(+)} & 1\textsuperscript{(+)}2\textsuperscript{(+)}4\textsuperscript{(+)} 6\textsuperscript{(+)} 8\textsuperscript{(+)} &   3\textsuperscript{(-)}5\textsuperscript{(-)} 7\textsuperscript{(-)}  & 1\textsuperscript{(+)}2\textsuperscript{(+)} 4\textsuperscript{(+)}6\textsuperscript{(+)} 8\textsuperscript{(+)} &   3\textsuperscript{(-)} 5\textsuperscript{(-)}7\textsuperscript{(-)}8\textsuperscript{(+)} & 1\textsuperscript{(+)}2\textsuperscript{(+)} 4\textsuperscript{(+)} 6\textsuperscript{(+)}8\textsuperscript{(+)} &  2\textsuperscript{(-)}3\textsuperscript{(-)} 5\textsuperscript{(-)}6\textsuperscript{(-)}7\textsuperscript{(-)}\\
 \cline{2-11} 
 & 1-1e-4 & mean &\textbf{2052.79} & 2947.92 & 13997.83 & \textbf{3095.06} & 15838.95 & \textbf{3302.84} & 8787.47 & \cellcolor{gray!25}\textbf{1234.72}\\
 & sd &705.46 & 427.45 & 2934.27 & 1923.73 & 3449.72 & 2019.49 & 252.18 & 192.41\\
\hline
\end{tabular}
\end{adjustbox}
\end{table*}
\begin{table*}[!hbt]
\centering
\caption{Table corresponding to the results Figure 3(d) with bsc type instances, with setting $r= 2000$ , $\nu = 50$, $\sigma_i^2 \in V_2$. The mean, standard deviation values and statistical tests of the partial offline error for GSEMO and MOEA/D variants for 2- and 3- formulations. 
}
\label{table:3d_b}
\begin{adjustbox}{max width=1.33\textheight}
\begin{tabular}{r|r|r|rr|rr|rr|rr}
\hline
\multirow{3}{*}{\rotatebox[origin=c]{90}{file-n}} & \multirow{3}{*}{$\alpha$} &\multirow{3}{*}{$stats$} & \multicolumn{2}{c}{$GSEMO$} & \multicolumn{2}{c}{$MOEA/D_{te}$} & \multicolumn{2}{c}{$MOEA/D_{pbi}$} & \multicolumn{2}{c}{$MOEA/D_{ws}$}\\
& & & $2D\textsuperscript{(1)}$ &  $3D\textsuperscript{(2)}$ & $2D\textsuperscript{(3)}$ & $3D\textsuperscript{(4)}$& $2D\textsuperscript{(5)}$ &$3D\textsuperscript{(6)}$ & $2D\textsuperscript{(7)}$ & $3D\textsuperscript{(8)}$ \\
\hline
\multirow{15}{*}{\rotatebox[origin=c]{90}{bsc-300}} & 1-1e-2 & mean &\cellcolor{gray!25}\textbf{4592.89} & 4914.08 & 76483.86 & \textbf{6718.80} & 75946.58 & \textbf{10617.98} & 34271.25 & \textbf{7464.36}\\
 & & sd &1323.39 & 1773.00 & 14242.59 & 3236.50 & 13866.86 & 3418.00 & 1744.87 & 3340.07\\
 & & stat & 3\textsuperscript{(-)} 5\textsuperscript{(-)}6\textsuperscript{(-)}7\textsuperscript{(-)}  &  3\textsuperscript{(-)} 5\textsuperscript{(-)}6\textsuperscript{(-)}7\textsuperscript{(-)}  & 1\textsuperscript{(+)}2\textsuperscript{(+)}4\textsuperscript{(+)} 6\textsuperscript{(+)} 8\textsuperscript{(+)} &   3\textsuperscript{(-)}5\textsuperscript{(-)} 7\textsuperscript{(-)}  & 1\textsuperscript{(+)}2\textsuperscript{(+)} 4\textsuperscript{(+)}6\textsuperscript{(+)} 8\textsuperscript{(+)} & 1\textsuperscript{(+)}2\textsuperscript{(+)}3\textsuperscript{(-)} 5\textsuperscript{(-)}   & 1\textsuperscript{(+)}2\textsuperscript{(+)} 4\textsuperscript{(+)}  8\textsuperscript{(+)} &   3\textsuperscript{(-)} 5\textsuperscript{(-)} 7\textsuperscript{(-)}\\
 \cline{2-11} 
 & 1-1e-4 & mean &\cellcolor{gray!25}\textbf{6428.00} & 6553.27 & 97643.48 & \textbf{7635.49} & 100364.57 & \textbf{10624.02} & 38213.63 & \textbf{8169.49}\\
 & & sd &2278.24 & 2407.54 & 11581.29 & 3501.58 & 13482.52 & 3412.78 & 2440.59 & 3539.14\\
\hline
\end{tabular}
\end{adjustbox}
\end{table*}

\begin{table*}[!hbt]
\centering
\caption{Table corresponding to the results Figure 3(d) with uncorr type instances, with setting $r= 2000$ , $\nu = 50$, $\sigma_i^2 \in V_2$. The mean, standard deviation values and statistical tests of the partial offline error for GSEMO and MOEA/D variants for 2- and 3- formulations. 
}

\label{table:3d_u}
\begin{adjustbox}{max width=1.35\textheight}
\begin{tabular}{r|r|r|rr|rr|rr|rr}
\hline
\multirow{3}{*}{\rotatebox[origin=c]{90}{file-n}} & \multirow{3}{*}{$\alpha$} &\multirow{3}{*}{$stats$} & \multicolumn{2}{c}{$GSEMO$} & \multicolumn{2}{c}{$MOEA/D_{te}$} & \multicolumn{2}{c}{$MOEA/D_{pbi}$} & \multicolumn{2}{c}{$MOEA/D_{ws}$}\\
& & & $2D\textsuperscript{(1)}$ &  $3D\textsuperscript{(2)}$ & $2D\textsuperscript{(3)}$ & $3D\textsuperscript{(4)}$& $2D\textsuperscript{(5)}$ &$3D\textsuperscript{(6)}$ & $2D\textsuperscript{(7)}$ & $3D\textsuperscript{(8)}$ \\
\hline
\multirow{15}{*}{\rotatebox[origin=c]{90}{uncorr-300}} &  1-1e-2 & mean &6394.49 & \cellcolor{gray!25}\textbf{5723.50} & 23774.28 & \textbf{7808.11} & 25178.10 & \textbf{16816.53} & 14269.16 & \textbf{6315.80}\\
 & & sd &1197.92 & 1283.46 & 2663.05 & 4009.47 & 3407.78 & 7300.67 & 1002.17 & 2073.14\\
 & & stat & 3\textsuperscript{(-)} 5\textsuperscript{(-)}6\textsuperscript{(-)}7\textsuperscript{(-)}  &  3\textsuperscript{(-)} 5\textsuperscript{(-)}6\textsuperscript{(-)}7\textsuperscript{(-)}  & 1\textsuperscript{(+)}2\textsuperscript{(+)}4\textsuperscript{(+)}   8\textsuperscript{(+)} &   3\textsuperscript{(-)}5\textsuperscript{(-)}6\textsuperscript{(-)}7\textsuperscript{(-)}  & 1\textsuperscript{(+)}2\textsuperscript{(+)} 4\textsuperscript{(+)} 7\textsuperscript{(+)}8\textsuperscript{(+)} & 1\textsuperscript{(+)}2\textsuperscript{(+)} 4\textsuperscript{(+)}  8\textsuperscript{(+)} & 1\textsuperscript{(+)}2\textsuperscript{(+)} 4\textsuperscript{(+)}5\textsuperscript{(-)} 8\textsuperscript{(+)} &   3\textsuperscript{(-)} 5\textsuperscript{(-)}6\textsuperscript{(-)}7\textsuperscript{(-)}\\
 \cline{2-11} 
 & 1-1e-4 & mean &7963.93 & \textbf{7820.36} & 28731.17 & \textbf{8904.89} & 29493.40 & \textbf{16929.70} & 16958.28 & \cellcolor{gray!25}\textbf{7777.07}\\
 & & sd &1908.03 & 1825.56 & 5432.41 & 3916.69 & 4675.16 & 7207.89 & 1686.93 & 2227.17\\
\hline
\end{tabular}
\end{adjustbox}
\end{table*}
\begin{table*}[!hbt]
\centering
\caption{Table corresponding to the results Figure 3(e) with bsc type instances, with setting $r= 2000$ , $\nu = 100$, $\sigma_i^2 \in V_1$. The mean, standard deviation values and statistical tests of the partial offline error for GSEMO and MOEA/D variants for 2- and 3- formulations. 
}
\label{table:3e_b}
\begin{adjustbox}{max width=1.35\textheight}
\begin{tabular}{r|r|r|rr|rr|rr|rr}
\hline
\multirow{3}{*}{\rotatebox[origin=c]{90}{file-n}} & \multirow{3}{*}{$\alpha$} &\multirow{3}{*}{$stats$} & \multicolumn{2}{c}{$GSEMO$} & \multicolumn{2}{c}{$MOEA/D_{te}$} & \multicolumn{2}{c}{$MOEA/D_{pbi}$} & \multicolumn{2}{c}{$MOEA/D_{ws}$}\\
& & & $2D\textsuperscript{(1)}$ &  $3D\textsuperscript{(2)}$ & $2D\textsuperscript{(3)}$ & $3D\textsuperscript{(4)}$& $2D\textsuperscript{(5)}$ &$3D\textsuperscript{(6)}$ & $2D\textsuperscript{(7)}$ & $3D\textsuperscript{(8)}$ \\
\hline
\multirow{15}{*}{\rotatebox[origin=c]{90}{bsc-300}} &  1-1e-2 & mean &\textbf{1185.44} & 1390.49 & 51656.91 & \cellcolor{gray!25}\textbf{767.31} & 50901.65 & \textbf{879.58} & 14113.42 & \textbf{1322.12}\\
 & & sd &291.86 & 207.48 & 9183.86 & 412.48 & 8969.17 & 332.18 & 213.10 & 336.51\\
 & & stat & 3\textsuperscript{(-)} 5\textsuperscript{(-)} 7\textsuperscript{(-)}  &  3\textsuperscript{(-)}4\textsuperscript{(+)}5\textsuperscript{(-)}6\textsuperscript{(+)}7\textsuperscript{(-)}  & 1\textsuperscript{(+)}2\textsuperscript{(+)}4\textsuperscript{(+)} 6\textsuperscript{(+)} 8\textsuperscript{(+)} &  2\textsuperscript{(-)}3\textsuperscript{(-)}5\textsuperscript{(-)} 7\textsuperscript{(-)}8\textsuperscript{(-)} & 1\textsuperscript{(+)}2\textsuperscript{(+)} 4\textsuperscript{(+)}6\textsuperscript{(+)} 8\textsuperscript{(+)} &  2\textsuperscript{(-)}3\textsuperscript{(-)} 5\textsuperscript{(-)}7\textsuperscript{(-)}  & 1\textsuperscript{(+)}2\textsuperscript{(+)} 4\textsuperscript{(+)} 6\textsuperscript{(+)}8\textsuperscript{(+)} &   3\textsuperscript{(-)}4\textsuperscript{(+)}5\textsuperscript{(-)} 7\textsuperscript{(-)}\\
 \cline{2-11} 
 & 1-1e-4 & mean &1728.74 & \textbf{1494.68} & 50233.25 & \cellcolor{gray!25}\textbf{905.95} & 52292.39 & \textbf{992.86} & 14280.62 & \textbf{1405.06}\\
 & & sd &2161.81 & 221.09 & 8857.26 & 414.82 & 12797.10 & 338.82 & 190.92 & 345.52\\
\hline
\end{tabular}
\end{adjustbox}
\end{table*}

\begin{table*}[!hbt]
\centering
\caption{Table corresponding to the results Figure 3(e) with uncorr type instances, with setting $r= 2000$ , $\nu = 100$, $\sigma_i^2 \in V_1$. The mean, standard deviation values and statistical tests of the partial offline error for GSEMO and MOEA/D variants for 2- and 3- formulations. 
}
\label{table:3e_u}
\begin{adjustbox}{max width=1.35\textheight}
\begin{tabular}{r|r|r|rr|rr|rr|rr}
\hline
\multirow{3}{*}{\rotatebox[origin=c]{90}{file-n}} & \multirow{3}{*}{$\alpha$} &\multirow{3}{*}{$stats$} & \multicolumn{2}{c}{$GSEMO$} & \multicolumn{2}{c}{$MOEA/D_{te}$} & \multicolumn{2}{c}{$MOEA/D_{pbi}$} & \multicolumn{2}{c}{$MOEA/D_{ws}$}\\
& & & $2D\textsuperscript{(1)}$ &  $3D\textsuperscript{(2)}$ & $2D\textsuperscript{(3)}$ & $3D\textsuperscript{(4)}$& $2D\textsuperscript{(5)}$ &$3D\textsuperscript{(6)}$ & $2D\textsuperscript{(7)}$ & $3D\textsuperscript{(8)}$ \\
\hline
\multirow{15}{*}{\rotatebox[origin=c]{90}{uncorr-300}} &  1-1e-2 & mean &\textbf{3043.37} & 3478.33 & 14905.19 & \textbf{3381.87} & 17292.78 & \textbf{3280.83} & 5726.79 & \cellcolor{gray!25}\textbf{1405.32}\\
 & & sd &892.65 & 447.33 & 2066.73 & 1515.03 & 3602.47 & 1220.65 & 408.31 & 180.95\\
 & & stat & 3\textsuperscript{(-)} 5\textsuperscript{(-)} 7\textsuperscript{(-)}8\textsuperscript{(+)} &  3\textsuperscript{(-)} 5\textsuperscript{(-)} 7\textsuperscript{(-)}8\textsuperscript{(+)} & 1\textsuperscript{(+)}2\textsuperscript{(+)}4\textsuperscript{(+)} 6\textsuperscript{(+)} 8\textsuperscript{(+)} &   3\textsuperscript{(-)}5\textsuperscript{(-)} 7\textsuperscript{(-)}8\textsuperscript{(+)} & 1\textsuperscript{(+)}2\textsuperscript{(+)} 4\textsuperscript{(+)}6\textsuperscript{(+)} 8\textsuperscript{(+)} &   3\textsuperscript{(-)} 5\textsuperscript{(-)}7\textsuperscript{(-)}8\textsuperscript{(+)} & 1\textsuperscript{(+)}2\textsuperscript{(+)} 4\textsuperscript{(+)} 6\textsuperscript{(+)}8\textsuperscript{(+)} & 1\textsuperscript{(-)}2\textsuperscript{(-)}3\textsuperscript{(-)}4\textsuperscript{(-)}5\textsuperscript{(-)}6\textsuperscript{(-)}7\textsuperscript{(-)}\\
 \cline{2-11} 
 & 1-1e-4 & mean &\textbf{2947.26} & 3646.04 & 15508.34 & \textbf{3632.17} & 16727.11 & \textbf{3528.85} & 5967.07 & \cellcolor{gray!25}\textbf{1572.59}\\
 & & sd &880.20 & 459.09 & 2541.57 & 1556.81 & 3005.51 & 1275.73 & 285.48 & 185.19\\
\hline

\end{tabular}
\end{adjustbox}
\end{table*}
\begin{table*}[!hbt]
\centering
\caption{Table corresponding to the results Figure 3(f) with bsc type instances, with setting $r= 2000$ , $\nu = 100$, $\sigma_i^2 \in V_2$. The mean, standard deviation values and statistical tests of the partial offline error for GSEMO and MOEA/D variants for 2- and 3- formulations. 
}

\label{table:3f_b}
\begin{adjustbox}{max width=1.35\textheight}
\begin{tabular}{r|r|r|rr|rr|rr|rr}
\hline
\multirow{3}{*}{\rotatebox[origin=c]{90}{file-n}} & \multirow{3}{*}{$\alpha$} &\multirow{3}{*}{$stats$} & \multicolumn{2}{c}{$GSEMO$} & \multicolumn{2}{c}{$MOEA/D_{te}$} & \multicolumn{2}{c}{$MOEA/D_{pbi}$} & \multicolumn{2}{c}{$MOEA/D_{ws}$}\\
& & & $2D\textsuperscript{(1)}$ &  $3D\textsuperscript{(2)}$ & $2D\textsuperscript{(3)}$ & $3D\textsuperscript{(4)}$& $2D\textsuperscript{(5)}$ &$3D\textsuperscript{(6)}$ & $2D\textsuperscript{(7)}$ & $3D\textsuperscript{(8)}$ \\
\hline
\multirow{15}{*}{\rotatebox[origin=c]{90}{bsc-300}} & 1-1e-2 & mean &10551.75 & \cellcolor{gray!25}\textbf{5535.05} & 79887.59 & \textbf{18374.61} & 83174.40 & \textbf{22884.14} & 20209.53 & \textbf{19680.15}\\
 & & sd &6879.48 & 1941.12 & 16150.97 & 6904.29 & 17665.42 & 7663.39 & 1676.78 & 6299.73\\
 & & stat & 3\textsuperscript{(-)} 5\textsuperscript{(-)}6\textsuperscript{(-)}7\textsuperscript{(-)}8\textsuperscript{(-)} &  3\textsuperscript{(-)}4\textsuperscript{(-)}5\textsuperscript{(-)}6\textsuperscript{(-)}7\textsuperscript{(-)}8\textsuperscript{(-)} & 1\textsuperscript{(+)}2\textsuperscript{(+)}4\textsuperscript{(+)} 6\textsuperscript{(+)}7\textsuperscript{(+)}8\textsuperscript{(+)} &  2\textsuperscript{(+)}3\textsuperscript{(-)}5\textsuperscript{(-)}    & 1\textsuperscript{(+)}2\textsuperscript{(+)} 4\textsuperscript{(+)}6\textsuperscript{(+)}7\textsuperscript{(+)}8\textsuperscript{(+)} & 1\textsuperscript{(+)}2\textsuperscript{(+)}3\textsuperscript{(-)} 5\textsuperscript{(-)}   & 1\textsuperscript{(+)}2\textsuperscript{(+)}3\textsuperscript{(-)} 5\textsuperscript{(-)}   & 1\textsuperscript{(+)}2\textsuperscript{(+)}3\textsuperscript{(-)} 5\textsuperscript{(-)}  \\
 \cline{2-11} 
 & 1-1e-4 & mean &16587.88 & \cellcolor{gray!25}\textbf{7437.16} & 97081.33 & \textbf{20192.04} & 97158.75 & \textbf{23887.48} & 22757.89 & \textbf{21771.35}\\
 & & sd &7614.45 & 2614.53 & 24875.67 & 7410.44 & 12864.74 & 8058.01 & 3087.86 & 6946.09\\
\hline
\end{tabular}
\end{adjustbox}
\end{table*}

\begin{table*}[!hbt]
\centering
\caption{Table corresponding to the results Figure 3(f) with uncorr type instances, with setting $r= 2000$ , $\nu = 100$, $\sigma_i^2 \in V_2$. The mean, standard deviation values and statistical tests of the partial offline error for GSEMO and MOEA/D variants for 2- and 3- formulations. 
}
\label{table:3f_u}
\begin{adjustbox}{max width=1.33\textheight}
\begin{tabular}{r|r|r|rr|rr|rr|rr}
\hline
\multirow{3}{*}{\rotatebox[origin=c]{90}{file-n}} & \multirow{3}{*}{$\alpha$} &\multirow{3}{*}{$stats$} & \multicolumn{2}{c}{$GSEMO$} & \multicolumn{2}{c}{$MOEA/D_{te}$} & \multicolumn{2}{c}{$MOEA/D_{pbi}$} & \multicolumn{2}{c}{$MOEA/D_{ws}$}\\
& & & $2D\textsuperscript{(1)}$ &  $3D\textsuperscript{(2)}$ & $2D\textsuperscript{(3)}$ & $3D\textsuperscript{(4)}$& $2D\textsuperscript{(5)}$ &$3D\textsuperscript{(6)}$ & $2D\textsuperscript{(7)}$ & $3D\textsuperscript{(8)}$ \\
\hline
\multirow{15}{*}{\rotatebox[origin=c]{90}{uncorr-300}} &  1-1e-2 & mean &8872.50 & \cellcolor{gray!25}\textbf{6830.50} & 26528.83 & \textbf{10586.87} & 26103.66 & \textbf{23631.82} & 10907.40 & \textbf{9757.16}\\
 & & sd &1870.02 & 1159.22 & 3309.30 & 6440.41 & 3288.55 & 6526.91 & 1204.51 & 2634.72\\
 & & stat & 3\textsuperscript{(-)} 5\textsuperscript{(-)}6\textsuperscript{(-)}   &  3\textsuperscript{(-)} 5\textsuperscript{(-)}6\textsuperscript{(-)}7\textsuperscript{(-)}  & 1\textsuperscript{(+)}2\textsuperscript{(+)}4\textsuperscript{(+)}  7\textsuperscript{(+)}8\textsuperscript{(+)} &   3\textsuperscript{(-)}5\textsuperscript{(-)}6\textsuperscript{(-)}   & 1\textsuperscript{(+)}2\textsuperscript{(+)} 4\textsuperscript{(+)} 7\textsuperscript{(+)}8\textsuperscript{(+)} & 1\textsuperscript{(+)}2\textsuperscript{(+)} 4\textsuperscript{(+)} 7\textsuperscript{(+)}8\textsuperscript{(+)} &  2\textsuperscript{(+)}3\textsuperscript{(-)} 5\textsuperscript{(-)}6\textsuperscript{(-)}  &   3\textsuperscript{(-)} 5\textsuperscript{(-)}6\textsuperscript{(-)} \\
 \cline{2-11} 
 & 1-1e-4 & mean &11744.26 & \cellcolor{gray!25}\textbf{9421.40} & 31709.75 & \textbf{11652.08} & 32162.31 & \textbf{23758.60} & 13598.55 & \textbf{11477.77}\\
 & & sd &2383.75 & 1544.52 & 4891.38 & 6338.20 & 4422.97 & 6488.55 & 1569.06 & 2896.42\\
\hline
\end{tabular}
\end{adjustbox}
\end{table*}
\end{landscape}

\end{document}